\documentclass[runningheads]{llncs}
\usepackage{graphicx}
\usepackage{color}

\usepackage[noend]{algpseudocode}
\usepackage{subfigure, multirow}
\usepackage{hyperref}
\usepackage{booktabs}
\usepackage{ulem}
\usepackage{array}
\usepackage{amsmath}
\usepackage{booktabs}
\usepackage{diagbox}
\usepackage{xspace}
\usepackage{cleveref}
\usepackage{amssymb}
\usepackage{bm}
\usepackage{makecell}

\usepackage{graphicx}
\usepackage{floatrow}
\usepackage{url}
\usepackage{setspace}
\newfloatcommand{capbtabbox}{table}[][\FBwidth]
\usepackage{pgfplots}
\usepackage{wrapfig}
\usepackage{ragged2e}
\usepackage{xcolor}
\pgfplotsset{compat=1.14}

\usepackage{bm}

\DeclareMathOperator*{\argmin}{arg\,min}

\newcommand{\bx}{{\bm{x}}}

\newcommand{\by}{{\bm{y}}}

\newcommand{\btheta}{{\bm{\theta}}}

\newcommand{\mt}{\mathcal{T}}
\newcommand{\ms}{\mathcal{S}}

\newcommand{\ma}{\mathcal{A}}
\newcommand{\mb}{\mathcal{B}}

\newcommand{\mv}{\mathcal{V}}

\newcommand{\mS}{\ms}
\newcommand{\mT}{\mt}

\makeatletter
\DeclareRobustCommand\onedot{\futurelet\@let@token\@onedot}
\def\@onedot{\ifx\@let@token.\else.\null\fi\xspace}
\def\eg{e.g\onedot} 
\def\ie{i.e\onedot} 
 
\def\etc{etc\onedot} 
 
\def\etal{et al\onedot}
\def\aka{aka\ }

\renewcommand{\paragraph}{%
	\@startsection{paragraph}{4}{\z@}%
	{0.1em \@plus 0.5ex \@minus 0.2ex}{-1em}%
	{\normalsize\bf}%
}
\makeatother

\begin{document}
\setcounter{page}{124}
\hypersetup{hidelinks}


%
\title{DeepCore: A Comprehensive Library for Coreset Selection in Deep Learning}

%
%
\author{Chengcheng Guo\inst{1\dagger}\and Bo Zhao\inst{2\dagger}\and Yanbing Bai\inst{1\ddagger}}
\authorrunning{C. Guo et al.}
%
\institute{Center for Applied Statistics, School of Statistics, Renmin University of China \and
School of Informatics, The University of Edinburgh\\
$^{\dagger}$ Equal contribution, 
$^{\ddagger}$ Corresponding author\\
{chengchengguo@ruc.edu.cn, bo.zhao@ed.ac.uk, ybbai@ruc.edu.cn}
}

\maketitle              
{
\vspace*{-60mm}\hspace*{-14mm}
\href{https://arxiv.org/abs/2204.08499}{\includegraphics[width=12mm]{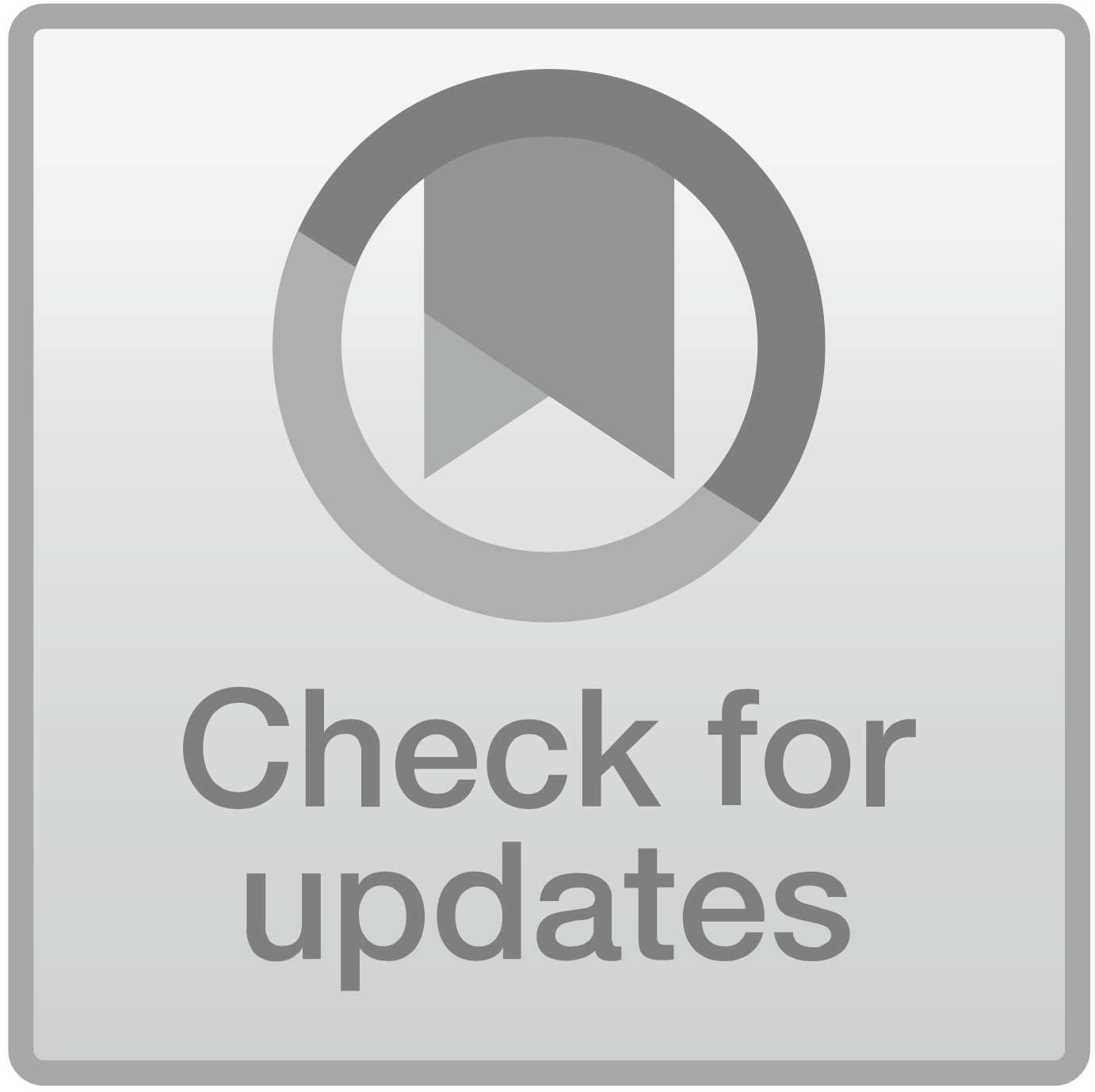}}
\vspace*{46mm}
}

\begin{abstract}
Coreset selection, which aims to select a subset of the most informative training samples, is a long-standing learning problem that can benefit many downstream tasks such as data-efficient learning, continual learning, neural architecture search, active learning, \etc. 
However, many existing coreset selection methods are not designed for deep learning, which may have high complexity and poor generalization performance. In addition, the recently proposed methods are evaluated on models, datasets, and settings of different complexities. 
To advance the research of coreset selection in deep learning, we contribute a comprehensive code library\footnote{The code is available in \url{https://github.com/PatrickZH/DeepCore}.\\}, namely \textit{DeepCore}, and provide an empirical study on popular coreset selection methods on CIFAR10 and ImageNet datasets.
Extensive experiments on CIFAR10 and ImageNet datasets verify that, although various methods have advantages in certain experiment settings, random selection is still a strong baseline.
\end{abstract}

\renewcommand{\thefootnote}{}
\footnote{\noindent\hspace*{-3.5mm} \copyright\ Springer Nature Switzerland AG 2022 \\
 \noindent\hspace*{-3.5mm} C. Strauss et al. (Eds.): DEXA 2022, LNCS 13108, pp. 124–138, 2022.\\
 \noindent\hspace*{-3.5mm} \textcolor{blue}{\url{https://doi.org/10.48550/arXiv.2204.08499}}}

\section{Introduction}
\label{sec:intro}

Deep learning has shown unprecedented success in many research areas such as computer vision, \etc. As it evolves, not only neural networks but also the training datasets are becoming increasingly larger, which requires massive memory and computation to achieve the state-of-the-art. 
One promising technique to reduce the computational cost is coreset selection \cite{mirzasoleiman2020coresets,killamsetty2021glister,paul2021deep,pmlr-v139-killamsetty21a} that aims to select a small subset of the most informative training samples $\ms$ from a given large training dataset $\mt$. The models trained on the coreset are supposed to have close generalization performance to those trained on the original training set.

Coreset selection has been widely studied since the era of traditional machine learning, whose research generally focuses on how to approximate the distribution of the whole dataset with a subset, for example, they assume that data are from a mixture of Gaussians in a given metric space \cite{welling2009herding,chen2010super,feldman2011scalable,bateni2014distributed,bachem2015coresets}. 
However, for those classic coreset selection methods proposed for traditional machine learning tasks, their effectiveness in deep learning is doubtful, due to the high computational complexity and fixed data representations.
Recently, the research of coreset selection for deep learning tasks emerges \cite{toneva2018empirical,paul2021deep,killamsetty2021glister}.
The newly developed coreset selection methods are evaluated in different settings in terms of models, datasets, tasks, and selection fractions, resulting in their performances hardly being compared fairly.

We focus our studies on image classification tasks. To address the above problems, in this paper, we provide an exhaustive empirical study on popular coreset selection methods in the same settings. 
We contribute a comprehensive code library, namely \textit{DeepCore}, for advancing the research of coreset selection in deep learning. Specifically, we re-implement 12 popular coreset selection methods in a unified framework based on PyTorch \cite{paszke2019pytorch}.
These methods are compared in settings of various selection fractions from 0.1\% to 90\% on CIFAR10 \cite{krizhevsky2009learning} and ImageNet-1K \cite{ILSVRC15} datasets. 
Besides the reported results in the paper, our library supports  popular deep neural architectures, image classification datasets and coreset selection settings.

\section{Review of Coreset Selection Methods}
\label{sec:survey}

In this section, we first formulate the problem of coreset selection. Then, brief surveys of methods and applications of coreset selection are provided respectively.

\subsection{Problem Statement}

In a learning task, we are given a large training set $\mt=\{(\bx_i, y_i)\}_{i=1}^{|\mt|}$, where $\bx_i\in\mathcal X$ is the input, $y_i\in\mathcal Y$ is the ground-truth label of $\bx_i$, where $\mathcal{X}$ and $\mathcal{Y}$ denote the input and output spaces, respectively. 
Coreset selection aims to find the most informative subset $\ms\subset \mt$ with the constraint $|\ms| < |\mt|$, so that the model $\btheta^\ms$ trained on $\ms$ has close generalization performance to the model $\btheta^\mt$ trained on the whole training set $\mt$. 

\subsection{Survey: Methodologies}\label{ssec:survey}
\subsubsection{Geometry Based Methods}
It is assumed that data points close to each other in the feature space tend to have similar properties. Therefore, geometry based methods \cite{chen2010super,sener2018active,sinha2020small,agarwal2020contextual} try to remove those data points providing redundant information then the left data points form a coreset $\ms$ where $|\ms|\ll|\mt|$.

{\textsc{Herding.}} The \textsc{Herding} method selects data points based on the distance between the coreset center and original dataset center in the feature space. The algorithm incrementally and greedily adds one sample each time into the coreset that can minimize distance between two centers \cite{welling2009herding,chen2010super}.

{\textsc{k-Center Greedy.}} This method tries to solves the \textit{minimax facility location} problem \cite{wolf2011facility}, \ie. selecting $k$ samples as $\ms$ from the full dataset $\mT$ such that the largest distance between a data point in $\mT\backslash \mS$ and its closest data point in $\ms$ is minimized:
\begin{equation}
\setlength{\abovedisplayskip}{4pt} \setlength{\belowdisplayskip}{4pt}
    \min_{\mS\subset \mt} \max_{\bx_i\in \mT\backslash \mS} \min_{\bx_j\in \mS} \mathcal{D}(\bx_i,\bx_j),
\end{equation}
where $\mathcal{D}(\cdot,\cdot)$ is the distance function. The problem is NP-hard, and a greedy approximation known as \textsc{k-Center Greedy} has been proposed in \cite{sener2018active}.  \textsc{k-Center Greedy} has been successfully extended to a wide range of applications, for instance, active learning \cite{sener2018active,agarwal2020contextual} and efficient GAN training \cite{sinha2020small}.

\subsubsection{Uncertainty Based Methods}
Samples with lower confidence may have a greater impact on model optimization than those with higher confidence, and should therefore be included in the coreset. The following are commonly used metrics of sample uncertainty given a certain classifier and training epoch, namely \textsc{Least Confidence}, \textsc{Entropy} and \textsc{Margin} \cite{coleman2019selection}, where $C$ is the number of classes. We select samples in descending order of the scores:
\begin{equation}
\setlength{\abovedisplayskip}{4pt} \setlength{\belowdisplayskip}{4pt}\label{eq:uncer}
\begin{aligned}
    s_{\textit{least confidence}}(\bx) = & 1-\max_{i=1,...,C} P(\hat y=i|\bx)\\
    s_{\textit{entropy}}(\bx) = & -\sum_{i=1}^{C} P(\hat y=i|\bx)\log P(\hat y=i|\bx)\\
    s_{\textit{margin}}(\bx) = & 1-\min_{y\neq \hat y}(P(\hat y|x)-P(y|x)).
\end{aligned}\end{equation}

\subsubsection{Error/Loss Based Methods}
In a dataset, training samples are more important if they contribute more to the error or loss when training neural networks.
Importance can be measured by the loss or gradient of each sample or its influence on other samples' prediction during model training. Those samples with the largest importance are selected as the coreset.


{\textsc{Forgetting Events}.} Toneva \etal \cite{toneva2018empirical} count how many times the \textit{forgetting} happens during the training, \ie  the misclassification of a sample in the current epoch after having been correctly classified in the previous epoch, formally $acc_i^{t} > acc_i^{t+1}$, where ${\rm acc}_i^t$ indicates the correctness (True or False) of the prediction of sample $i$ at epoch $t$. The number of forgetting reveals intrinsic properties of the training data, allowing for the removal of unforgettable examples with minimal performance drop. 

{\textsc{GraNd} and \textsc{EL2N} \textit{Scores.}}   The \textsc{GraNd} score \cite{paul2021deep} of  sample $(\bx, y)$ at epoch $t$ is defined as
\begin{equation}\setlength{\abovedisplayskip}{4pt} \setlength{\belowdisplayskip}{4pt}
    \chi_t(\bx,y)\triangleq\mathbb{E}_{\btheta_t}||\nabla_{\btheta_t}\ell(\bx,y; \btheta_t)||_2.
\end{equation} It measures the average contribution from each sample to the decline of the training loss at early epoch $t$ across several different independent runs.
The score calculated at early training stages, \eg after a few epochs, works well, thus this method requires less computational cost.
An approximation of the \textsc{GraNd} score is also provided, named \textsc{EL2N} score, which measures the norm of error vector:
\begin{equation}\setlength{\abovedisplayskip}{4pt} \setlength{\belowdisplayskip}{4pt}
    \chi_t^*(\bx,y)\triangleq\mathbb{E}_{\btheta_t}||p(\btheta_t,\bx)-\by||_2.
\end{equation} 
 
{\textsc{Importance Sampling}.}
In importance sampling (or adaptive sampling),  we define $s(\bx, y)$ is the upper-bounded (worst-case) contribution to the total loss function from the data point $(\bx, y)$, \aka  sensitivity score. It can be formulated as: \begin{equation}\setlength{\abovedisplayskip}{4pt} \setlength{\belowdisplayskip}{4pt}
    s(\bx, y)=\max_{\btheta\in\btheta}\frac{\ell(\bx, y; \btheta)}{\sum_{(\bx', y')\in \mT}\ell(\bx', y'; \btheta)},
\end{equation}
where $\ell(\bx, y)$ is a non-negative cost function with  parameter $\btheta\in\btheta$.
For each data point in $\mT$, the probability of being selected is set as $p(\bx, y)=\frac{s(\bx, y)}{\sum_{(\bx, y)\in\mt} s(\bx, y)}$. The coreset $\mS$ is constructed based on the probabilities \cite{bachem2015coresets,munteanu2018coresets}. Similar ideas are proposed in \textit{Black box learners} \cite{dasgupta2019teaching} and \textsc{Jtt} \cite{liu2021just}, where wrongly classified samples will be upweighted or their sampling probability will be increased.  

\subsubsection{Decision Boundary Based Methods} 
Since data points distributed near the decision boundary are hard to separate, those data points closest to the decision boundary can also be used as the coreset.
 
{\textsc{Adversarial DeepFool}.} 
While exact distance to the decision boundary is inaccessible,  Ducoffe and Precioso \cite{ducoffe2018adversarial} seek the approximation of these distances in the input space $\mathcal{X}$. By giving perturbations to samples until the predictive labels of samples are changed, those data points require the smallest adversarial perturbation are closest to the decision boundary.
 
{\textsc{Contrastive Active Learning}.} To find data points near the decision boundary, Contrastive Active Learning (\textsc{Cal}) \cite{DBLP:journals/corr/abs-2109-03764} selects samples whose predictive likelihood diverges the most from their neighbors to construct the coreset.

\subsubsection{Gradient Matching Based Methods}
Deep models are usually trained using (stochastic) gradient descent algorithm. Therefore, we expect that the gradients produced by the full training dataset $\sum_{(\bx, y)\in \mT}\nabla_\btheta\ell(\bx, y; \btheta)$ can be replaced by the (weighted) gradients produced by a subset $\sum_{(\bx, y)\in \mS}w_\bx\nabla_\btheta \ell(\bx, y; \btheta)$ with minimal difference:
\begin{equation}
\setlength{\abovedisplayskip}{4pt}
\setlength{\belowdisplayskip}{4pt}
\begin{aligned}
\min_{\mathbf{w},\mS}  \mathcal{D}(\frac1{|\mT|}\sum\limits_{(\bx, y)\in \mT}\nabla_\btheta\ell(\bx, y; \btheta), \frac1{|\mathbf{w}|_1}\sum\limits_{(\bx, y)\in \mS} w_\bx\nabla_\btheta \ell(\bx, y; \btheta)) \\
\quad   s.t. \quad  \mS\subset \mT, \;  w_\bx\geq0,
\end{aligned}
\end{equation} 
where $\mathbf{w}$ is the subset weight vector, $|\mathbf{w}|_1$ is the sum of the absolute values and $\mathcal{D}(\cdot,\cdot)$ measures the distance between two gradients.

{\textsc{Craig.}} Mirzasoleiman \etal \cite{mirzasoleiman2020coresets} try to find an optimal coreset that approximates the full dataset gradients under a maximum error  $\varepsilon$  by converting gradient matching problem to the maximization of a monotone submodular function $F$ and then use greedy approach to optimize  $F$.

{\textsc{GradMatch.}} Compared to \textsc{Craig}, the \textsc{GradMatch} \cite{pmlr-v139-killamsetty21a} method is able to achieve the same error $\varepsilon$ of the gradient matching but with a smaller subset. \textsc{GradMatch} introduces a squared l2 regularization term over the weight vector $\mathbf{w}$ with coefficient $\lambda$ to discourage assigning large weights to individual samples. To solve the optimization problem, it presents a greedy algorithm -- \textit{Orthogonal Matching Pursuit}, which can guarantee $ 1-exp(\frac{-\lambda}{\lambda+k\nabla^2_{max}})$ error with the constraint $|\mS|\leq k$, $k$ is a preset constant.

\subsubsection{Bilevel Optimization Based Methods}
Coreset selection can be posed as a bilevel optimization problem.
Existing studies usually consider the selection of subset (optimization of samples $\ms$ or selection weights $\mathbf{w}$) as the outer objective and the optimization of model parameters $\btheta$ on $\ms$ as the inner objective. 
Representative methods include cardinality-constrained bilevel optimization \cite{borsos2020coresets} for continual learning,  \textsc{Retrieve} for semi-supervised learning (SSL) \cite{killamsetty2021retrieve}, and \textsc{Glister} \cite{killamsetty2021glister} for supervised learning and active learning.  


{\textsc{Retrieve}.} The \textsc{Retrieve} method \cite{killamsetty2021retrieve} discusses the scenario of SSL under bilevel optimization, where we have both a labeled set $\mt$ and an unlabled set $\mathcal{P}$. The bilevel optimization problem in \textsc{Retrieve} is formulated as 
\begin{equation}\label{eq:retrieve}
\begin{aligned}
    \mathbf{w}^{*}=\argmin_\mathbf{w} \sum_{(\bx, y)\in \mT}\ell_{s}(\bx, y; \argmin_\btheta (\sum_{(\bx, y)\in \mT}\ell_s(\bx, y; \btheta)+\lambda \sum_{\bx \in \mathcal{P}} w_x\ell_{u}(\bx; \btheta))),
    \end{aligned}
\end{equation}
where $\ell_s$ is the labeled-data loss, \eg cross-entropy and $\ell_u$ is the unlabeled-data loss for SSL, \eg consistency-regularization loss. $\lambda$ is the regularization coefficient.

{\textsc{Glister.}} To guarantee the robustness, \textsc{Glister} \cite{killamsetty2021glister}  introduces a validation set $\mv$ on the outer optimization and the log-likelihood $\ell\ell$ in the bilevel optimization:

\begin{equation}\setlength{\abovedisplayskip}{4pt} \setlength{\belowdisplayskip}{4pt}
\mS^*=\mathop{\arg\max}\limits_{\mS\subset \mt}\sum\limits _{(\bx, y)\in \mv}\ell\ell(\bx, y; \mathop{\arg\max}\limits_\btheta\sum\limits_{(\bx, y)\in \mS}\ell\ell(\bx, y; \btheta)).
\end{equation}

\subsubsection{Submodularity Based Methods}
Submodular functions \cite{iyer2013submodular} are set functions $f: 2^\mv\rightarrow \mathbb{R}$, which return a real value for any $\mathcal{U}\subset \mv$. $f$ is a submodular function, if for $\ma\subset\mb\subset \mv$ and $\forall x\in \mv\backslash\mb$: \begin{equation}\setlength{\abovedisplayskip}{4pt} \setlength{\belowdisplayskip}{4pt}
    f(\ma\cup\{x\})-f(\ma)\geq f(\mb\cup\{x\})-f(\mb).
\end{equation} Submodular functions naturally measure the diversity and information, thus can be a powerful tool for coreset selection by maximizing them. 
Many functions obey the above definition, \eg   {Graph Cut} (GC), {Facility Location} (FL), Log Determinant \cite{iyer2021submodular}, \etc. 
For maximizing submodular functions under cardinality constraint, greedy algorithms have been proved to have a bounded approximation factor of $1-\frac 1 e$ \cite{nemhauser1978analysis}.


{\textsc{Fass}.} Wei \etal \cite{wei2015submodularity} discuss the connection between likelihood functions and submodularity, proving that under a cardinality constraint, maximizing likelihood function is equivalent to maximization of submodular functions for Naïve Bayes or Nearest Neighbor classifier, naturally providing a powerful tool for coreset selection. By introducing submodularity into  Naive Bayes and Nearest Neighbor, they propose a novel framework for active learning namely \textsc{Filtered Active Submodular Selection (Fass)}.

{\textsc{Prism}.} Kaushal \etal \cite{kaushal2021prism} develop \textsc{Prism}, a submodular method for \textit{targeted subset selection}, which is a learning scenario similar to active learning. In targeted subset selection, a subset $\ms$ will be selected to be labeled from a large unlabeled set $\mathcal{P}$, with additional requirement that $\ms$ has to be aligned with the targeted set $\mt$ of specific user intent.

{\textsc{Similar}.} Kothawade \etal \cite{kothawade2021similar} introduce \textsc{Similar}, a unified framework of submodular methods that successfully extends submodularity to broader settings which may involve rare classes, redundancy, out-of-distribution data, \etc.

\subsubsection{Proxy Based Methods}
Many coreset selection methods require to train models on the whole dataset for calculating features or some metrics for one or many times.
To reduce this training cost, \textsc{Selection via Proxy} methods \cite{coleman2019selection,sachdeva2021svp} are proposed, which train a lighter or shallower version of the target models as proxy models. Specifically, they create proxy models by reducing hidden layers, narrowing dimensions, or cutting down training epochs. Then, coresets are selected more efficiently on these proxy models.

\subsection{Survey: Applications}
\paragraph{Data-efficient Learning.} The basic application of coreset selection is to enable efficient machine learning \cite{toneva2018empirical,mirzasoleiman2020coresets,paul2021deep,pmlr-v139-killamsetty21a}. Training models on coresets can reduce the training cost while preserving testing performance. Especially, in Neural Architecture Search (NAS) \cite{shim2021core}, thousands to millions deep models have to be trained and then evaluated on the same dataset. Coreset can be used as a proxy dataset to efficiently train and evaluate candidates \cite{coleman2019selection,sachdeva2021svp}, which significantly reduces computational cost.

\paragraph{Continual Learning.} Coreset selection is also a key technique to construct memory for continual learning or incremental learning \cite{aljundi2019gradient,borsos2020coresets,yoon2021online}, in order to relieve the catastrophic forgetting problem. In the popular continual learning setting, a memory buffer is maintained to store informative training samples from previous tasks for rehearsal in future tasks. It is proven that continual learning performance heavily relies on the quality of memory, \ie coreset \cite{knoblauch2020optimal}.

\paragraph{Active Learning.} Active learning \cite{settles2009active,settles2011theories} aims to achieve better performance with the minimal query cost by selecting informative samples from the unlabeled pool $\mathcal{P}$ to label. Thus, it can be posed as a coreset selection problem \cite{wei2015submodularity,sener2018active,ducoffe2018adversarial,kothawade2021similar,DBLP:journals/corr/abs-2109-03764}.

Besides the above, coreset selection is studied and successfully applied in many other machine learning problems, such as robust learning against noise \cite{NEURIPS2020_8493eeac,killamsetty2021retrieve,kothawade2021similar}, clustering \cite{bateni2014distributed,bachem2015coresets,sohler2018strong}, semi-supervised learning \cite{borsos2021semi,killamsetty2021retrieve}, unsupervised learning \cite{ju2021extending}, efficient GAN training \cite{sinha2020small}, regression tasks \cite{munteanu2018coresets,chhaya2020coresets} \etc.

\section{DeepCore Library}
\label{sec:lib}

In the literature, coreset selection methods have been proposed and tested in different experiment settings in terms of dataset, model architecture, coreset size, augmentation, training strategy, \etc. This may lead to unfair comparisons between different methods and unconvincing conclusions. For instance, some methods may have only been evaluated on MNIST with shallow models, while others are tested on the challenging ImageNet dataset with deep neural networks. Even though tested on the same dataset, different works are likely to use different training strategies and data augmentations which significantly affect the performance. Furthermore, it causes future researchers inconvenience in identifying and improving the state-of-the-art.

Therefore, we develop \textit{DeepCore}, an extensive and extendable code library, for coreset selection in deep learning, reproducing dozens of popular and advanced coreset selection methods and enabling a fair comparison of different methods in the same experimental settings. DeepCore is highly modular, allowing to add new architectures, datasets, methods and learning scenarios easily. We build DeepCore on PyTorch \cite{paszke2019pytorch}.   

\paragraph{Coreset Methods.} 
We list the methods that have been re-implemented in DeepCore according to the categories in \ref{ssec:survey}, they are 1) geometry based methods \textsc{Contextual Diversity} (CD) \cite{agarwal2020contextual}, \textsc{Herding} \cite{welling2009herding} and \textsc{k-Center Greedy} \cite{sener2018active}; 2) uncertainty based methods \textsc{Least Confidence}, \textsc{Entropy} and \textsc{Margin}  \cite{coleman2019selection}; 3) error/loss based methods \textsc{Forgetting} \cite{toneva2018empirical} and \textsc{GraNd} \cite{paul2021deep}; 4) decision boundary based methods \textsc{Cal} \cite{DBLP:journals/corr/abs-2109-03764} and \textsc{DeepFool} \cite{ducoffe2018adversarial}; 5) gradient matching based methods \textsc{Craig} \cite{mirzasoleiman2020coresets} and \textsc{GradMatch} \cite{pmlr-v139-killamsetty21a}; 6) bilevel optimization methods \textsc{Glister} \cite{killamsetty2021glister}; and 7) submodularity based methods with \textsc{Graph Cut} (GC) and \textsc{Facility Location} (FL) functions \cite{iyer2021submodular}. We also have \textsc{Random} selection as the baseline.

\paragraph{Datasets.} We provide the experiment results on CIFAR10 \cite{krizhevsky2009learning} and ImageNet-1K \cite{ILSVRC15} in this paper. Besides, our DeepCore has provided the interface for other popular computer vision datasets, namely MNIST \cite{lecun1998gradient}, QMNIST \cite{yadav2019cold}, FashionMNIST \cite{xiao2017fashion}, SVHN \cite{netzer2011reading}, CIFAR100 \cite{krizhevsky2009learning} and TinyImageNet \cite{le2015tiny}.

\paragraph{Network Architectures.} We provide the code of popular architectures, namely MLP, LeNet \cite{lecun1989backpropagation}, AlexNet \cite{NIPS2012_c399862d}, VGG \cite{simonyan2014very}, Inception-v3 \cite{szegedy2016rethinking}, ResNet \cite{he2016deep}, WideResNet \cite{zagoruyko2016wide} and MobileNet-v3 \cite{DBLP:journals/corr/abs-1905-02244}.

\section{Experiment Results}
\label{sec:exp}
In this section, we use our DeepCore to evaluate different coreset selection methods in multiple learning settings on CIFAR10 and ImageNet-1K datasets. ResNet-18 is used as the default architecture in all experiments.

\begin{table}[]

  \centering
  \caption{Coreset selection performances on CIFAR10. We train randomly initialized ResNet-18 on the coresets of CIFAR10 produced by different methods and then test on the real testing set.}  \label{tab:cifar}
  \resizebox{\linewidth}{!}{
 \footnotesize{
 \setlength{\tabcolsep}{2pt}
\begin{tabular}{ccccccccccccc}
\midrule
Fraction     & 0.1\%       & 0.5\%       & 1\%       & 5\%       & 10\%      & 20\%      & 30\%      & 40\%      & 50\%      & 60\%      & 90\%      & 100\%        \\ \cmidrule(lr){2-13}\cmidrule(lr){1-1}
Random         & 21.0\hspace{0.02em}$\pm$\hspace{0.02em}0.3          & 30.8\hspace{0.02em}$\pm$\hspace{0.02em}0.6          & 36.7\hspace{0.02em}$\pm$\hspace{0.02em}1.7          & 64.5\hspace{0.02em}$\pm$\hspace{0.02em}1.1          & 75.7\hspace{0.02em}$\pm$\hspace{0.02em}2.0          & \textbf{87.1\hspace{0.02em}$\pm$\hspace{0.02em}0.5}          & 90.2\hspace{0.02em}$\pm$\hspace{0.02em}0.3          & 92.1\hspace{0.02em}$\pm$\hspace{0.02em}0.1          & 93.3\hspace{0.02em}$\pm$\hspace{0.02em}0.2          & 94.0\hspace{0.02em}$\pm$\hspace{0.02em}0.2          & 95.2\hspace{0.02em}$\pm$\hspace{0.02em}0.1          & 95.6\hspace{0.02em}$\pm$\hspace{0.02em}0.1 \\ \cmidrule(lr){2-13}
CD   \cite{agarwal2020contextual}   & 15.8\hspace{0.02em}$\pm$\hspace{0.02em}1.2          & 20.5\hspace{0.02em}$\pm$\hspace{0.02em}0.7          & 23.6\hspace{0.02em}$\pm$\hspace{0.02em}1.9          & 38.1\hspace{0.02em}$\pm$\hspace{0.02em}2.2          & 58.8\hspace{0.02em}$\pm$\hspace{0.02em}2.0          & 81.3\hspace{0.02em}$\pm$\hspace{0.02em}2.5          & 90.8\hspace{0.02em}$\pm$\hspace{0.02em}0.5          & 93.3\hspace{0.02em}$\pm$\hspace{0.02em}0.4          & 94.3\hspace{0.02em}$\pm$\hspace{0.02em}0.2          & 94.6\hspace{0.02em}$\pm$\hspace{0.02em}0.6          & 95.4\hspace{0.02em}$\pm$\hspace{0.02em}0.1          & 95.6\hspace{0.02em}$\pm$\hspace{0.02em}0.1 \\
Herding   \cite{welling2009herding}  & 20.2\hspace{0.02em}$\pm$\hspace{0.02em}2.3          & 27.3\hspace{0.02em}$\pm$\hspace{0.02em}1.5          & 34.8\hspace{0.02em}$\pm$\hspace{0.02em}3.3          & 51.0\hspace{0.02em}$\pm$\hspace{0.02em}3.1          & 63.5\hspace{0.02em}$\pm$\hspace{0.02em}3.4          & 74.1\hspace{0.02em}$\pm$\hspace{0.02em}2.5          & 80.1\hspace{0.02em}$\pm$\hspace{0.02em}2.2          & 85.2\hspace{0.02em}$\pm$\hspace{0.02em}0.9          & 88.0\hspace{0.02em}$\pm$\hspace{0.02em}1.1          & 89.8\hspace{0.02em}$\pm$\hspace{0.02em}0.9          & 94.6\hspace{0.02em}$\pm$\hspace{0.02em}0.4          & 95.6\hspace{0.02em}$\pm$\hspace{0.02em}0.1 \\
k-Center Greedy \cite{sener2018active} & 18.5\hspace{0.02em}$\pm$\hspace{0.02em}0.3          & 26.8\hspace{0.02em}$\pm$\hspace{0.02em}1.2          & 31.1\hspace{0.02em}$\pm$\hspace{0.02em}1.2          & 51.4\hspace{0.02em}$\pm$\hspace{0.02em}2.1          & 75.8\hspace{0.02em}$\pm$\hspace{0.02em}2.4          & 87.0\hspace{0.02em}$\pm$\hspace{0.02em}0.3          & 90.9\hspace{0.02em}$\pm$\hspace{0.02em}0.4          & 92.8\hspace{0.02em}$\pm$\hspace{0.02em}0.1          & 93.9\hspace{0.02em}$\pm$\hspace{0.02em}0.2          & 94.1\hspace{0.02em}$\pm$\hspace{0.02em}0.1          & 95.4\hspace{0.02em}$\pm$\hspace{0.02em}0.1          & 95.6\hspace{0.02em}$\pm$\hspace{0.02em}0.1 \\ \cmidrule(lr){2-13}
Least Confidence \cite{coleman2019selection} & 14.2\hspace{0.02em}$\pm$\hspace{0.02em}0.9          & 17.2\hspace{0.02em}$\pm$\hspace{0.02em}1.8          & 19.8\hspace{0.02em}$\pm$\hspace{0.02em}2.2          & 36.2\hspace{0.02em}$\pm$\hspace{0.02em}1.9          & 57.6\hspace{0.02em}$\pm$\hspace{0.02em}3.1          & 81.9\hspace{0.02em}$\pm$\hspace{0.02em}2.2          & 90.3\hspace{0.02em}$\pm$\hspace{0.02em}0.4          & 93.1\hspace{0.02em}$\pm$\hspace{0.02em}0.5          & 94.5\hspace{0.02em}$\pm$\hspace{0.02em}0.1          & 94.7\hspace{0.02em}$\pm$\hspace{0.02em}0.1          & 95.5\hspace{0.02em}$\pm$\hspace{0.02em}0.1          & 95.6\hspace{0.02em}$\pm$\hspace{0.02em}0.1 \\
Entropy   \cite{coleman2019selection}   & 14.6\hspace{0.02em}$\pm$\hspace{0.02em}2.2          & 17.5\hspace{0.02em}$\pm$\hspace{0.02em}1.3          & 21.1\hspace{0.02em}$\pm$\hspace{0.02em}1.3          & 35.3\hspace{0.02em}$\pm$\hspace{0.02em}3.0          & 57.6\hspace{0.02em}$\pm$\hspace{0.02em}2.8          & 81.9\hspace{0.02em}$\pm$\hspace{0.02em}0.4          & 89.8\hspace{0.02em}$\pm$\hspace{0.02em}1.6          & 93.2\hspace{0.02em}$\pm$\hspace{0.02em}0.2          & 94.4\hspace{0.02em}$\pm$\hspace{0.02em}0.3          & \textbf{95.0\hspace{0.02em}$\pm$\hspace{0.02em}0.1}          & 95.4\hspace{0.02em}$\pm$\hspace{0.02em}0.1          & 95.6\hspace{0.02em}$\pm$\hspace{0.02em}0.1 \\
Margin   \cite{coleman2019selection}     & 17.2\hspace{0.02em}$\pm$\hspace{0.02em}1.1          & 21.7\hspace{0.02em}$\pm$\hspace{0.02em}1.6          & 28.2\hspace{0.02em}$\pm$\hspace{0.02em}1.0          & 43.4\hspace{0.02em}$\pm$\hspace{0.02em}3.3          & 59.9\hspace{0.02em}$\pm$\hspace{0.02em}2.9          & 81.7\hspace{0.02em}$\pm$\hspace{0.02em}3.2          & 90.9\hspace{0.02em}$\pm$\hspace{0.02em}0.4          & 93.0\hspace{0.02em}$\pm$\hspace{0.02em}0.2          & 94.3\hspace{0.02em}$\pm$\hspace{0.02em}0.3          & 94.8\hspace{0.02em}$\pm$\hspace{0.02em}0.3          & 95.5\hspace{0.02em}$\pm$\hspace{0.02em}0.1          & 95.6\hspace{0.02em}$\pm$\hspace{0.02em}0.1 \\ \cmidrule(lr){2-13}
Forgetting   \cite{toneva2018empirical}  & 21.4\hspace{0.02em}$\pm$\hspace{0.02em}0.5 & 29.8\hspace{0.02em}$\pm$\hspace{0.02em}1.0 & 35.2\hspace{0.02em}$\pm$\hspace{0.02em}1.6 & 52.1\hspace{0.02em}$\pm$\hspace{0.02em}2.2 & 67.0\hspace{0.02em}$\pm$\hspace{0.02em}1.5 & 86.6\hspace{0.02em}$\pm$\hspace{0.02em}0.6 & \textbf{91.7\hspace{0.02em}$\pm$\hspace{0.02em}0.3} & 93.5\hspace{0.02em}$\pm$\hspace{0.02em}0.2 & 94.1\hspace{0.02em}$\pm$\hspace{0.02em}0.1 & 94.6\hspace{0.02em}$\pm$\hspace{0.02em}0.2 & 95.3\hspace{0.02em}$\pm$\hspace{0.02em}0.1& 95.6\hspace{0.02em}$\pm$\hspace{0.02em}0.1 \\
GraNd  \cite{paul2021deep}   & 17.7\hspace{0.02em}$\pm$\hspace{0.02em}1.0 & 24.0\hspace{0.02em}$\pm$\hspace{0.02em}1.1 & 26.7\hspace{0.02em}$\pm$\hspace{0.02em}1.3 & 39.8\hspace{0.02em}$\pm$\hspace{0.02em}2.3 & 52.7\hspace{0.02em}$\pm$\hspace{0.02em}1.9 & 78.2\hspace{0.02em}$\pm$\hspace{0.02em}2.9 & 91.2\hspace{0.02em}$\pm$\hspace{0.02em}0.7 & \textbf{93.7\hspace{0.02em}$\pm$\hspace{0.02em}0.3} & \textbf{94.6\hspace{0.02em}$\pm$\hspace{0.02em}0.1} & \textbf{95.0\hspace{0.02em}$\pm$\hspace{0.02em}0.2} & 95.5\hspace{0.02em}$\pm$\hspace{0.02em}0.2          & 95.6\hspace{0.02em}$\pm$\hspace{0.02em}0.1 \\ \cmidrule(lr){2-13}
Cal   \cite{DBLP:journals/corr/abs-2109-03764}      & 22.7\hspace{0.02em}$\pm$\hspace{0.02em}2.7          & 33.1\hspace{0.02em}$\pm$\hspace{0.02em}2.3          & 37.8\hspace{0.02em}$\pm$\hspace{0.02em}2.0          & 60.0\hspace{0.02em}$\pm$\hspace{0.02em}1.4          & 71.8\hspace{0.02em}$\pm$\hspace{0.02em}1.0          & 80.9\hspace{0.02em}$\pm$\hspace{0.02em}1.1          & 86.0\hspace{0.02em}$\pm$\hspace{0.02em}1.9          & 87.5\hspace{0.02em}$\pm$\hspace{0.02em}0.8          & 89.4\hspace{0.02em}$\pm$\hspace{0.02em}0.6          & 91.6\hspace{0.02em}$\pm$\hspace{0.02em}0.9          & 94.7\hspace{0.02em}$\pm$\hspace{0.02em}0.3          & 95.6\hspace{0.02em}$\pm$\hspace{0.02em}0.1 \\
DeepFool     \cite{ducoffe2018adversarial}    & 17.6\hspace{0.02em}$\pm$\hspace{0.02em}0.4          & 22.4\hspace{0.02em}$\pm$\hspace{0.02em}0.8          & 27.6\hspace{0.02em}$\pm$\hspace{0.02em}2.2          & 42.6\hspace{0.02em}$\pm$\hspace{0.02em}3.5          & 60.8\hspace{0.02em}$\pm$\hspace{0.02em}2.5          & 83.0\hspace{0.02em}$\pm$\hspace{0.02em}2.3          & 90.0\hspace{0.02em}$\pm$\hspace{0.02em}0.7          & 93.1\hspace{0.02em}$\pm$\hspace{0.02em}0.2          & 94.1\hspace{0.02em}$\pm$\hspace{0.02em}0.1          & 94.8\hspace{0.02em}$\pm$\hspace{0.02em}0.2          & 95.5\hspace{0.02em}$\pm$\hspace{0.02em}0.1          & 95.6\hspace{0.02em}$\pm$\hspace{0.02em}0.1 \\ \cmidrule(lr){2-13}
Craig     \cite{mirzasoleiman2020coresets}   & 22.5\hspace{0.02em}$\pm$\hspace{0.02em}1.2          & 27.0\hspace{0.02em}$\pm$\hspace{0.02em}0.7          & 31.7\hspace{0.02em}$\pm$\hspace{0.02em}1.1          & 45.2\hspace{0.02em}$\pm$\hspace{0.02em}2.9          & 60.2\hspace{0.02em}$\pm$\hspace{0.02em}4.4          & 79.6\hspace{0.02em}$\pm$\hspace{0.02em}3.1          & 88.4\hspace{0.02em}$\pm$\hspace{0.02em}0.5          & 90.8\hspace{0.02em}$\pm$\hspace{0.02em}1.4          & 93.3\hspace{0.02em}$\pm$\hspace{0.02em}0.6          & 94.2\hspace{0.02em}$\pm$\hspace{0.02em}0.2          & 95.5\hspace{0.02em}$\pm$\hspace{0.02em}0.1          & 95.6\hspace{0.02em}$\pm$\hspace{0.02em}0.1 \\
GradMatch   \cite{pmlr-v139-killamsetty21a}  & 17.4\hspace{0.02em}$\pm$\hspace{0.02em}1.7          & 25.6\hspace{0.02em}$\pm$\hspace{0.02em}2.6          & 30.8\hspace{0.02em}$\pm$\hspace{0.02em}1.0          & 47.2\hspace{0.02em}$\pm$\hspace{0.02em}0.7          & 61.5\hspace{0.02em}$\pm$\hspace{0.02em}2.4          & 79.9\hspace{0.02em}$\pm$\hspace{0.02em}2.6          & 87.4\hspace{0.02em}$\pm$\hspace{0.02em}2.0          & 90.4\hspace{0.02em}$\pm$\hspace{0.02em}1.5          & 92.9\hspace{0.02em}$\pm$\hspace{0.02em}0.6          & 93.2\hspace{0.02em}$\pm$\hspace{0.02em}1.0          & 93.7\hspace{0.02em}$\pm$\hspace{0.02em}0.5          & 95.6\hspace{0.02em}$\pm$\hspace{0.02em}0.1 \\ \cmidrule(lr){2-13}
Glister  \cite{killamsetty2021glister}  & 19.5\hspace{0.02em}$\pm$\hspace{0.02em}2.1          & 27.5\hspace{0.02em}$\pm$\hspace{0.02em}1.4          & 32.9\hspace{0.02em}$\pm$\hspace{0.02em}2.4          & 50.7\hspace{0.02em}$\pm$\hspace{0.02em}1.5          & 66.3\hspace{0.02em}$\pm$\hspace{0.02em}3.5          & 84.8\hspace{0.02em}$\pm$\hspace{0.02em}0.9          & 90.9\hspace{0.02em}$\pm$\hspace{0.02em}0.3          & 93.0\hspace{0.02em}$\pm$\hspace{0.02em}0.2          & 94.0\hspace{0.02em}$\pm$\hspace{0.02em}0.3          & 94.8\hspace{0.02em}$\pm$\hspace{0.02em}0.2          & \textbf{95.6\hspace{0.02em}$\pm$\hspace{0.02em}0.2} & 95.6\hspace{0.02em}$\pm$\hspace{0.02em}0.1 \\ \cmidrule(lr){2-13}
FL     \cite{iyer2021submodular}    & 22.3\hspace{0.02em}$\pm$\hspace{0.02em}2.0          & 31.6\hspace{0.02em}$\pm$\hspace{0.02em}0.6          & 38.9\hspace{0.02em}$\pm$\hspace{0.02em}1.4          & 60.8\hspace{0.02em}$\pm$\hspace{0.02em}2.5          & 74.7\hspace{0.02em}$\pm$\hspace{0.02em}1.3          & 85.6\hspace{0.02em}$\pm$\hspace{0.02em}1.9          & 91.4\hspace{0.02em}$\pm$\hspace{0.02em}0.4 & 93.2\hspace{0.02em}$\pm$\hspace{0.02em}0.3          & 93.9\hspace{0.02em}$\pm$\hspace{0.02em}0.2          & 94.5\hspace{0.02em}$\pm$\hspace{0.02em}0.3          & 95.5\hspace{0.02em}$\pm$\hspace{0.02em}0.2          & 95.6\hspace{0.02em}$\pm$\hspace{0.02em}0.1 \\
GC   \cite{iyer2021submodular}    &   \textbf{24.3\hspace{0.02em}$\pm$\hspace{0.02em}1.5} & \textbf{34.9\hspace{0.02em}$\pm$\hspace{0.02em}2.3} & \textbf{42.8\hspace{0.02em}$\pm$\hspace{0.02em}1.3} & \textbf{65.7\hspace{0.02em}$\pm$\hspace{0.02em}1.2} & \textbf{76.6\hspace{0.02em}$\pm$\hspace{0.02em}1.5} & 84.0\hspace{0.02em}$\pm$\hspace{0.02em}0.5          & 87.8\hspace{0.02em}$\pm$\hspace{0.02em}0.4          & 90.6\hspace{0.02em}$\pm$\hspace{0.02em}0.3          & 93.2\hspace{0.02em}$\pm$\hspace{0.02em}0.3          & 94.4\hspace{0.02em}$\pm$\hspace{0.02em}0.3          & 95.4\hspace{0.02em}$\pm$\hspace{0.02em}0.1          & 95.6\hspace{0.02em}$\pm$\hspace{0.02em}0.1\\  \midrule
\end{tabular}
}
}

\end{table}%
\subsection{CIFAR10 Results}
\label{ssec:cifar}

\begin{figure*}[h]
  \centering
  \includegraphics[width=1\linewidth]{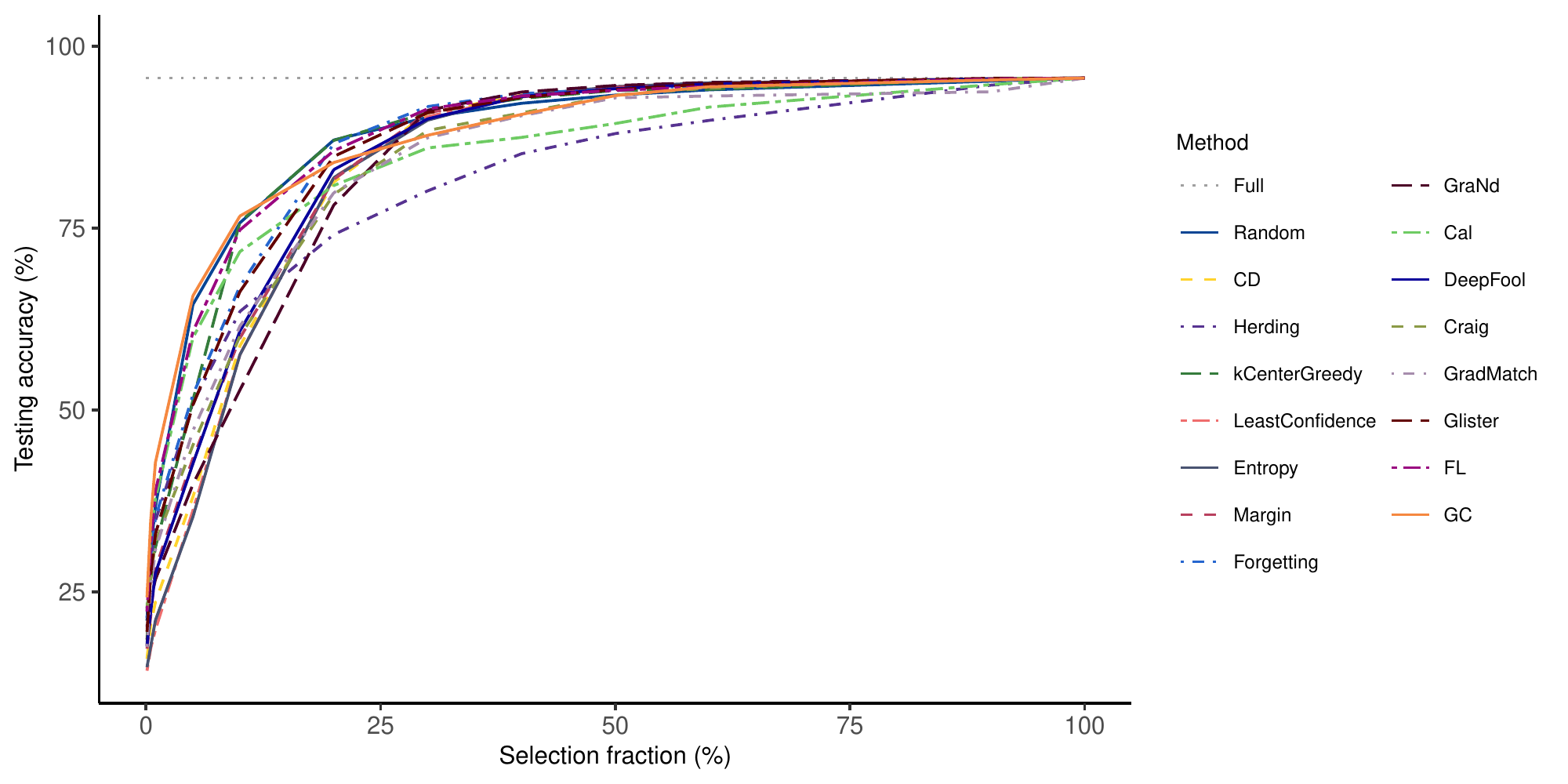}
\caption{Coreset selection performances in curves on CIFAR10. We train randomly initialized ResNet-18 on the coresets of CIFAR10 produced by different methods and then test on the real testing set. Detailed numbers are provided in Tab. \ref{tab:cifar}.}
\label{fig:cifar}
\end{figure*}

For CIFAR10 experiments, we use SGD as the optimizer with batch size 128, initial learning rate 0.1, Cosine decay scheduler, momentum 0.9, weight decay $5\times10^{-4}$ and 200 training epochs. We select subsets with fractions of 0.1\%, 0.5\%, 1\%, 5\%, 10\%, 20\%, 30\%, 40\%, 50\%, 60\%, 90\% of the whole training set respectively. 
The training on the whole dataset can be considered as the upper-bound. For data augmentation, we apply random crop with 4-pixel padding and random flipping on the 32$\times$32 training images.

For some methods, the gradient, prediction probability, or feature vector of each sample is required to implement sample selection.
For a fair comparison, we use the ResNet-18 models trained on the whole dataset for 10 epochs to extract above-mentioned metrics.
When gradient vector $\nabla_\theta\ell(\bx,y;\theta)$ is required, we use the gradients of the parameters in the final fully-connected layer as suggested in many previous studies \cite{mirzasoleiman2020coresets,pmlr-v139-killamsetty21a,killamsetty2021glister}.
This allows gradient vectors to be easily obtained without back-propagation throughout the whole network.
While DeepCore supports both balanced and imbalance sample selection, experiments in this paper all adopt balanced selection, namely, the same number of samples are selected for every class.

Tab. \ref{tab:cifar} shows the detailed results of different methods on CIFAR10, and Fig. \ref{fig:cifar} depicts the performance curves. The mean and standard deviation is calculated with 5 random seeds. 
Good experimental results come from the submodular function based methods, in both small and large learning setting. Especially in small fractions of 0.1\%-1\%, the advantage of submodular function based methods is obvious.
Graph Cut (GC) is more prominent among them, and achieves the best results when selecting 0.1\% to 10\% of the training data.
In particular, Graph Cut outperforms the other methods by more than 5\% in the testing accuracy when 50 samples are selected per class, \ie 1\% of the whole training set.
\textsc{Cal} also shows superiority in small fractions between 0.1\%-5\%, with performance comparable to Facility Location (FL). However, its superiority disappears when the coreset size increases, especially when selecting more than 30\% training data.
Except the above methods, all other methods fail to outperform the random sampling baseline in small settings between 0.1\% and 1\%.
Forgetting method outperforms others in 30\%-fraction setting. 
Between 40\% and 60\%, \textsc{GraNd} and uncertainty score based methods stand out. 
In all fraction settings, \textsc{GradMatch} and \textsc{Herding} barely beat the random sampling.
For \textsc{GradMatch}, the experiment setting in the original paper is adaptive sampling, where subsets iteratively updated along with network training. Here, for a fair comparison, coresets are selected and then fixed for all training epochs. \textsc{Herding} is originally designed for fixed representations from a mixture of Gaussians, thus its performance heavily depends on the embedding function. 
Note that the above findings are based on one hyper-parameter setting, the findings may change if hyper-parameters change. For example, \textsc{Herding} may have better performances if the model for feature extraction is fully trained. We study the influence of some hyper-parameters later.

\subsection{ImageNet Results}

\begin{table}[htbp]

  \centering
  \caption{Coreset selection performances on ImageNet-1K. We train randomly initialized ResNet-18 on the coresets of ImageNet produced by different methods and then test on the real testing set.}\label{tab:imagenet}
   \resizebox{0.98\linewidth}{!}
  {
 \scriptsize{
 \setlength{\tabcolsep}{2pt}
\begin{tabular}{cccccccc}
\midrule
                   & 0.1\%         & 0.5\%         &  1\%         & 5\%         & 10\%       & 30\%  & 100\%         \\ \cmidrule(lr){1-1}\cmidrule(lr){2-8}
Random       & 0.76\hspace{0.02em}$\pm$\hspace{0.02em}0.01 & 3.78\hspace{0.02em}$\pm$\hspace{0.02em}0.14 & 8.85\hspace{0.02em}$\pm$\hspace{0.02em}0.46 & 40.09\hspace{0.02em}$\pm$\hspace{0.02em}0.21 & 52.10\hspace{0.02em}$\pm$\hspace{0.02em}0.22 &
\textbf{64.11\hspace{0.02em}$\pm$\hspace{0.02em}0.05}
& 69.52\hspace{0.02em}$\pm$\hspace{0.02em}0.45 \\\cmidrule(lr){2-8}
CD               & -    & 1.18\hspace{0.02em}$\pm$\hspace{0.02em}0.06 & 2.16\hspace{0.02em}$\pm$\hspace{0.02em}0.18 & 25.82\hspace{0.02em}$\pm$\hspace{0.02em}2.02 & 43.84\hspace{0.02em}$\pm$\hspace{0.02em}0.12 &
62.13\hspace{0.02em}$\pm$\hspace{0.02em}0.45
&69.52\hspace{0.02em}$\pm$\hspace{0.02em}0.45 \\
Herding          & 0.34\hspace{0.02em}$\pm$\hspace{0.02em}0.01 & 1.70\hspace{0.02em}$\pm$\hspace{0.02em}0.13 & 4.17\hspace{0.02em}$\pm$\hspace{0.02em}0.26 & 17.41\hspace{0.02em}$\pm$\hspace{0.02em}0.34 & 28.06\hspace{0.02em}$\pm$\hspace{0.02em}0.05 & 48.58\hspace{0.02em}$\pm$\hspace{0.02em}0.49 & 69.52\hspace{0.02em}$\pm$\hspace{0.02em}0.45 \\
k-Center Greedy    & -    & 1.57\hspace{0.02em}$\pm$\hspace{0.02em}0.09 & 2.96\hspace{0.02em}$\pm$\hspace{0.02em}0.24 & 27.36\hspace{0.02em}$\pm$\hspace{0.02em}0.08 & 44.84\hspace{0.02em}$\pm$\hspace{0.02em}1.03 &
62.12\hspace{0.02em}$\pm$\hspace{0.02em}0.46 & 69.52\hspace{0.02em}$\pm$\hspace{0.02em}0.45 \\\cmidrule(lr){2-8}
Least Confidence & 0.29\hspace{0.02em}$\pm$\hspace{0.02em}0.04 & 1.03\hspace{0.02em}$\pm$\hspace{0.02em}0.25 & 2.05\hspace{0.02em}$\pm$\hspace{0.02em}0.38 & 27.05\hspace{0.02em}$\pm$\hspace{0.02em}3.25 & 44.47\hspace{0.02em}$\pm$\hspace{0.02em}1.42 & 61.80\hspace{0.02em}$\pm$\hspace{0.02em}0.33 & 69.52\hspace{0.02em}$\pm$\hspace{0.02em}0.45 \\
Entropy          & 0.31\hspace{0.02em}$\pm$\hspace{0.02em}0.02 & 1.01\hspace{0.02em}$\pm$\hspace{0.02em}0.17 & 2.26\hspace{0.02em}$\pm$\hspace{0.02em}0.30 & 28.21\hspace{0.02em}$\pm$\hspace{0.02em}2.83 & 44.68\hspace{0.02em}$\pm$\hspace{0.02em}1.54 &
61.82\hspace{0.02em}$\pm$\hspace{0.02em}0.31&69.52\hspace{0.02em}$\pm$\hspace{0.02em}0.45 \\
Margin           & 0.47\hspace{0.02em}$\pm$\hspace{0.02em}0.02 & 1.99\hspace{0.02em}$\pm$\hspace{0.02em}0.29 & 4.73\hspace{0.02em}$\pm$\hspace{0.02em}0.64 & 35.99\hspace{0.02em}$\pm$\hspace{0.02em}1.67 & 50.29\hspace{0.02em}$\pm$\hspace{0.02em}0.92 & 63.62\hspace{0.02em}$\pm$\hspace{0.02em}0.15& 69.52\hspace{0.02em}$\pm$\hspace{0.02em}0.45 \\\cmidrule(lr){2-8}
Forgetting       & 0.76\hspace{0.02em}$\pm$\hspace{0.02em}0.01 & 4.69\hspace{0.02em}$\pm$\hspace{0.02em}0.17 & 14.02\hspace{0.02em}$\pm$\hspace{0.02em}0.13 & \textbf{47.64\hspace{0.02em}$\pm$\hspace{0.02em}0.03} & \textbf{55.12\hspace{0.02em}$\pm$\hspace{0.02em}0.13} & 62.49\hspace{0.02em}$\pm$\hspace{0.02em}0.11 & 69.52\hspace{0.02em}$\pm$\hspace{0.02em}0.45 \\
GraNd   & 
1.04\hspace{0.02em}$\pm$\hspace{0.02em}0.04 & 7.02\hspace{0.02em}$\pm$\hspace{0.02em}0.05 & \textbf{18.10\hspace{0.02em}$\pm$\hspace{0.02em}0.22} & 43.53\hspace{0.02em}$\pm$\hspace{0.02em}0.19 & 49.92\hspace{0.02em}$\pm$\hspace{0.02em}0.21 &
57.98\hspace{0.02em}$\pm$\hspace{0.02em}0.17&69.52\hspace{0.02em}$\pm$\hspace{0.02em}0.45\\\cmidrule(lr){2-8}
Cal              & \textbf{1.29\hspace{0.02em}$\pm$\hspace{0.02em}0.09} & 7.50\hspace{0.02em}$\pm$\hspace{0.02em}0.26 & 15.94\hspace{0.02em}$\pm$\hspace{0.02em}1.30 & 38.32\hspace{0.02em}$\pm$\hspace{0.02em}0.78 & 46.49\hspace{0.02em}$\pm$\hspace{0.02em}0.29 & 58.31\hspace{0.02em}$\pm$\hspace{0.02em}0.32& 69.52\hspace{0.02em}$\pm$\hspace{0.02em}0.45 \\\cmidrule(lr){2-8}
Craig            & 1.13\hspace{0.02em}$\pm$\hspace{0.02em}0.08 & 5.44\hspace{0.02em}$\pm$\hspace{0.02em}0.52 & 9.40\hspace{0.02em}$\pm$\hspace{0.02em}1.69 & 32.30\hspace{0.02em}$\pm$\hspace{0.02em}1.24 & 38.77\hspace{0.02em}$\pm$\hspace{0.02em}0.56 & 44.89\hspace{0.02em}$\pm$\hspace{0.02em}3.72& 69.52\hspace{0.02em}$\pm$\hspace{0.02em}0.45 \\
GradMatch        & 0.93\hspace{0.02em}$\pm$\hspace{0.02em}0.04  & 5.20\hspace{0.02em}$\pm$\hspace{0.02em}0.22 & 12.28\hspace{0.02em}$\pm$\hspace{0.02em}0.49 & 40.16\hspace{0.02em}$\pm$\hspace{0.02em}2.28 & 45.91\hspace{0.02em}$\pm$\hspace{0.02em}1.73 & 52.69\hspace{0.02em}$\pm$\hspace{0.02em}2.16& 69.52\hspace{0.02em}$\pm$\hspace{0.02em}0.45 \\\cmidrule(lr){2-8}
Glister          & 0.98\hspace{0.02em}$\pm$\hspace{0.02em}0.06 & 5.91\hspace{0.02em}$\pm$\hspace{0.02em}0.42 & 14.87\hspace{0.02em}$\pm$\hspace{0.02em}0.14 & 44.95\hspace{0.02em}$\pm$\hspace{0.02em}0.28 & 52.04\hspace{0.02em}$\pm$\hspace{0.02em}1.18 &60.26\hspace{0.02em}$\pm$\hspace{0.02em}0.28& 69.52\hspace{0.02em}$\pm$\hspace{0.02em}0.45 \\\cmidrule(lr){2-8}
FL               & 1.23\hspace{0.02em}$\pm$\hspace{0.02em}0.03 & 5.78\hspace{0.02em}$\pm$\hspace{0.02em}0.08 & 12.72\hspace{0.02em}$\pm$\hspace{0.02em}0.21 & 40.85\hspace{0.02em}$\pm$\hspace{0.02em}1.25 & 51.05\hspace{0.02em}$\pm$\hspace{0.02em}0.59 &  63.14\hspace{0.02em}$\pm$\hspace{0.02em}0.03 &69.52\hspace{0.02em}$\pm$\hspace{0.02em}0.45 \\
GC               & 1.21\hspace{0.02em}$\pm$\hspace{0.02em}0.09 & \textbf{7.66\hspace{0.02em}$\pm$\hspace{0.02em}0.43} & 16.43\hspace{0.02em}$\pm$\hspace{0.02em}0.53 & 42.23\hspace{0.02em}$\pm$\hspace{0.02em}0.60 & 50.53\hspace{0.02em}$\pm$\hspace{0.02em}0.42 & 63.22\hspace{0.02em}$\pm$\hspace{0.02em}0.26& 69.52\hspace{0.02em}$\pm$\hspace{0.02em}0.45 \\  \midrule
\end{tabular}
}
}

\end{table}%

For ImageNet, we train ResNet-18 models on coresets with batch size 256 for 200 epochs.
The training images are randomly cropped and then resized to 224$\times$224. The left-right flipping with the probability of 0.5 is also implemented. 
Other experimental settings and hyper-parameters are consistent with CIFAR10 experiments. Due to the long running time of \textsc{DeepFool} on ImageNet, its results are not provided. For \textsc{k-Center Greedy} and \textsc{Contextual Diversity}, here we do not provide the results when only 1 sample is selected from each class (i.e. fraction of 0.1\%), because their first sample is drawn randomly from each class as initialization. Hence, they are identical to \textsc{Random} baseline for fraction 0.1\% on ImageNet. We run all experiments for 3 times with random seeds.

Experiment results are given in Tab. \ref{tab:imagenet}. The results show that error based methods, \textsc{Forgetting} and \textsc{GraNd}, generally have better performance on ImageNet. Especially, \textsc{Forgetting} overwhelms \textsc{Random} when fewer than 10\% data are selected as the coreset. However, none of methods will outperform \textsc{Random} when the coreset size is large, i.e. 30\% data. \textsc{Random} is still a strong and stable baseline.
The same to that on CIFAR10, these findings on ImageNet may vary for different hyper-parameters.

\subsection{Cross-architecture Generalization}\label{ssec:cross}
We conduct cross-architecture experiments to examine whether methods with good performance are model-agnostic, \ie, whether coresets perform  well when being selected on one architecture and then tested on other architectures.
We do experiments on four representative methods, (\textsc{Forgetting}, \textsc{Glister}, \textsc{GraNd} and \textsc{Graph Cut}) with
four representative architectures (VGG-16 \cite{simonyan2014very}, Inception-v3 \cite{szegedy2016rethinking}, ResNet-18 \cite{he2016deep} and WideResNet-16-8 \cite{zagoruyko2016wide})
under two selection fractions (1\% and 10\%). All other unspecified settings are the same to those in Sec. \ref{ssec:cifar}. 
In Tab. \ref{tab:cross}, the rows represent models used to obtain coresets, and the columns indicate models on which coresets are evaluated. We can see submodular selection with Graph Cut provides stably good testing results, regardless of which model architecture is used to perform the selection. 
However, \textsc{GraNd} shows preference of the model on which gradient norms are computed. Coresets obtained on Inception-v3 generally have the best performance, while those obtained on ResNet-18 are the worst. The possible reason is that the ranking of gradient norm is sensitive to the architecture.
The architecture used to implement selection also has obvious influence on \textsc{Glister} and \textsc{Forgetting} methods.

\begin{table}[htbp]
  \centering
  \caption{Cross-architecture generalization performance (\%) of four representative methods (\textsc{Forgetting}, \textsc{GraNd}, \textsc{Glister} and \textsc{Graph Cut}). The coreset is selected based on one (row) architecture and then evaluated on another (column) architecture.}
     \scriptsize
 \resizebox{1\linewidth}{!}
  {
  \begin{tabular}{ccccccccc}
    \toprule
    C\textbackslash{}T & VGG-16 & Inception-v3 & ResNet-18 & WRN-16-8 & VGG-16 & Inception-v3 & ResNet-18 & WRN-16-8 \\
    \midrule
    \textbf{Random} & \multicolumn{4}{c}{\textbf{1\%}}      & \multicolumn{4}{c}{\textbf{10\%}} \\ \cmidrule(lr){1-1}\cmidrule(lr){2-5}\cmidrule(lr){6-9}
    Random Selection & 15.36\hspace{0.02em}$\pm$\hspace{0.02em}2.03 & 32.98\hspace{0.02em}$\pm$\hspace{0.02em}1.20 & 36.74\hspace{0.02em}$\pm$\hspace{0.02em}1.69 & 45.77\hspace{0.02em}$\pm$\hspace{0.02em}1.17 & 78.03\hspace{0.02em}$\pm$\hspace{0.02em}0.92 & 76.01\hspace{0.02em}$\pm$\hspace{0.02em}0.82 & 75.72\hspace{0.02em}$\pm$\hspace{0.02em}2.02 & 82.72\hspace{0.02em}$\pm$\hspace{0.02em}0.54 \\
    \midrule
    \textbf{Forgetting} & \multicolumn{4}{c}{\textbf{1\%}}      & \multicolumn{4}{c}{\textbf{10\%}} \\
    \cmidrule(lr){1-1}\cmidrule(lr){2-5}\cmidrule(lr){6-9}
    VGG-16 & 17.56\hspace{0.02em}$\pm$\hspace{0.02em}3.42 & 31.37\hspace{0.02em}$\pm$\hspace{0.02em}0.63 & 35.07\hspace{0.02em}$\pm$\hspace{0.02em}1.38 & 40.30\hspace{0.02em}$\pm$\hspace{0.02em}1.94 & 72.71\hspace{0.02em}$\pm$\hspace{0.02em}2.26 & 70.68\hspace{0.02em}$\pm$\hspace{0.02em}1.85 & \textbf{71.53\hspace{0.02em}$\pm$\hspace{0.02em}0.42} & 80.71\hspace{0.02em}$\pm$\hspace{0.02em}1.11 \\
Inception-v3 & 21.81\hspace{0.02em}$\pm$\hspace{0.02em}3.04 & \textbf{33.27\hspace{0.02em}$\pm$\hspace{0.02em}1.70} & \textbf{36.94\hspace{0.02em}$\pm$\hspace{0.02em}1.28} & \textbf{41.52\hspace{0.02em}$\pm$\hspace{0.02em}1.71} & \textbf{72.94\hspace{0.02em}$\pm$\hspace{0.02em}0.63} & \textbf{71.15\hspace{0.02em}$\pm$\hspace{0.02em}2.84} & 70.40\hspace{0.02em}$\pm$\hspace{0.02em}2.09 & \textbf{81.51\hspace{0.02em}$\pm$\hspace{0.02em}0.95} \\
ResNet18 & \textbf{22.81\hspace{0.02em}$\pm$\hspace{0.02em}3.46} & 32.64\hspace{0.02em}$\pm$\hspace{0.02em}1.33 & 35.20\hspace{0.02em}$\pm$\hspace{0.02em}1.59 & 39.45\hspace{0.02em}$\pm$\hspace{0.02em}0.62 & 70.87\hspace{0.02em}$\pm$\hspace{0.02em}1.27 & 66.87\hspace{0.02em}$\pm$\hspace{0.02em}1.82 & 66.99\hspace{0.02em}$\pm$\hspace{0.02em}1.48 & 79.19\hspace{0.02em}$\pm$\hspace{0.02em}0.38 \\
WRN-16-8 & 20.53\hspace{0.02em}$\pm$\hspace{0.02em}3.49 & 28.46\hspace{0.02em}$\pm$\hspace{0.02em}1.48 & 31.79\hspace{0.02em}$\pm$\hspace{0.02em}1.11 & 35.92\hspace{0.02em}$\pm$\hspace{0.02em}1.97 & 67.68\hspace{0.02em}$\pm$\hspace{0.02em}1.37 & 64.38\hspace{0.02em}$\pm$\hspace{0.02em}1.82 & 65.59\hspace{0.02em}$\pm$\hspace{0.02em}2.03 & 75.59\hspace{0.02em}$\pm$\hspace{0.02em}1.09 \\
    \midrule
    \textbf{GraNd} & \multicolumn{4}{c}{\textbf{1\%}}      & \multicolumn{4}{c}{\textbf{10\%}} \\
    \cmidrule(lr){1-1}\cmidrule(lr){2-5}\cmidrule(lr){6-9}
    VGG-16          & \textbf{18.61\hspace{0.02em}$\pm$\hspace{0.02em}3.84} & 29.78\hspace{0.02em}$\pm$\hspace{0.02em}0.90  & 33.77\hspace{0.02em}$\pm$\hspace{0.02em}0.87 & 38.07\hspace{0.02em}$\pm$\hspace{0.02em}1.75 & 69.74\hspace{0.02em}$\pm$\hspace{0.02em}1.48 & 65.90\hspace{0.02em}$\pm$\hspace{0.02em}1.88  & 65.45\hspace{0.02em}$\pm$\hspace{0.02em}1.33 & 76.63\hspace{0.02em}$\pm$\hspace{0.02em}0.74 \\
Inception-v3    & 15.94\hspace{0.02em}$\pm$\hspace{0.02em}2.50  & \textbf{31.46\hspace{0.02em}$\pm$\hspace{0.02em}0.98} & \textbf{34.73\hspace{0.02em}$\pm$\hspace{0.02em}1.04} & \textbf{40.16\hspace{0.02em}$\pm$\hspace{0.02em}1.83} & \textbf{73.51\hspace{0.02em}$\pm$\hspace{0.02em}0.75} & \textbf{70.52\hspace{0.02em}$\pm$\hspace{0.02em}3.15} & \textbf{70.07\hspace{0.02em}$\pm$\hspace{0.02em}2.91} & \textbf{79.62\hspace{0.02em}$\pm$\hspace{0.02em}1.27} \\
ResNet18        & 14.42\hspace{0.02em}$\pm$\hspace{0.02em}3.10  & 25.91\hspace{0.02em}$\pm$\hspace{0.02em}1.59 & 26.69\hspace{0.02em}$\pm$\hspace{0.02em}1.30 & 30.40\hspace{0.02em}$\pm$\hspace{0.02em}0.75  & 61.05\hspace{0.02em}$\pm$\hspace{0.02em}1.91 & 58.48\hspace{0.02em}$\pm$\hspace{0.02em}3.95 & 52.73\hspace{0.02em}$\pm$\hspace{0.02em}1.86 & 70.96\hspace{0.02em}$\pm$\hspace{0.02em}1.14 \\
WRN-16-8 & 14.59\hspace{0.02em}$\pm$\hspace{0.02em}4.03 & 28.68\hspace{0.02em}$\pm$\hspace{0.02em}1.43 & 32.30\hspace{0.02em}$\pm$\hspace{0.02em}1.87  & 35.88\hspace{0.02em}$\pm$\hspace{0.02em}3.18 & 61.49\hspace{0.02em}$\pm$\hspace{0.02em}1.81 & 57.19\hspace{0.02em}$\pm$\hspace{0.02em}2.42 & 57.82\hspace{0.02em}$\pm$\hspace{0.02em}2.27 & 69.19\hspace{0.02em}$\pm$\hspace{0.02em}1.92\\

    \midrule
    \textbf{Glister} & \multicolumn{4}{c}{\textbf{1\%}}      & \multicolumn{4}{c}{\textbf{10\%}} \\
    \cmidrule(lr){1-1}\cmidrule(lr){2-5}\cmidrule(lr){6-9}
    VGG-16 & 14.5\hspace{0.02em}$\pm$\hspace{0.02em}3.86 & 31.08\hspace{0.02em}$\pm$\hspace{0.02em}2.30 & 34.10\hspace{0.02em}$\pm$\hspace{0.02em}1.71 & 39.45\hspace{0.02em}$\pm$\hspace{0.02em}2.55 & 71.71\hspace{0.02em}$\pm$\hspace{0.02em}1.83 & 70.23\hspace{0.02em}$\pm$\hspace{0.02em}1.78 & 69.31\hspace{0.02em}$\pm$\hspace{0.02em}2.19 & 77.74\hspace{0.02em}$\pm$\hspace{0.02em}0.68 \\
    Inception-v3 & \textbf{19.74\hspace{0.02em}$\pm$\hspace{0.02em}4.01} & \textbf{32.05\hspace{0.02em}$\pm$\hspace{0.02em}1.12} & \textbf{35.52\hspace{0.02em}$\pm$\hspace{0.02em}2.09} & \textbf{41.24\hspace{0.02em}$\pm$\hspace{0.02em}1.39} & \textbf{73.15\hspace{0.02em}$\pm$\hspace{0.02em}1.94} & \textbf{71.32\hspace{0.02em}$\pm$\hspace{0.02em}1.77} & \textbf{71.03\hspace{0.02em}$\pm$\hspace{0.02em}1.39} & \textbf{78.57\hspace{0.02em}$\pm$\hspace{0.02em}1.45} \\
    ResNet-18 & 15.16\hspace{0.02em}$\pm$\hspace{0.02em}4.47 & 30.41\hspace{0.02em}$\pm$\hspace{0.02em}2.08 & 32.93\hspace{0.02em}$\pm$\hspace{0.02em}2.36 & 37.64\hspace{0.02em}$\pm$\hspace{0.02em}1.83 & 67.37\hspace{0.02em}$\pm$\hspace{0.02em}2.48 & 66.34\hspace{0.02em}$\pm$\hspace{0.02em}2.18 & 66.26\hspace{0.02em}$\pm$\hspace{0.02em}3.47 & 75.36\hspace{0.02em}$\pm$\hspace{0.02em}1.52 \\
    WRN-16-8 & 14.16\hspace{0.02em}$\pm$\hspace{0.02em}4.15 & 28.39\hspace{0.02em}$\pm$\hspace{0.02em}2.50 & 32.83\hspace{0.02em}$\pm$\hspace{0.02em}0.98 & 37.05\hspace{0.02em}$\pm$\hspace{0.02em}2.72 & 70.70\hspace{0.02em}$\pm$\hspace{0.02em}2.40 & 64.25\hspace{0.02em}$\pm$\hspace{0.02em}2.53 & 66.88\hspace{0.02em}$\pm$\hspace{0.02em}2.97 & 75.07\hspace{0.02em}$\pm$\hspace{0.02em}2.96 \\
    \midrule
    \textbf{Graph Cut} & \multicolumn{4}{c}{\textbf{1\%}}      & \multicolumn{4}{c}{\textbf{10\%}} \\
    \cmidrule(lr){1-1}\cmidrule(lr){2-5}\cmidrule(lr){6-9}
    VGG-16 & 27.47\hspace{0.02em}$\pm$\hspace{0.02em}4.00 & 37.38\hspace{0.02em}$\pm$\hspace{0.02em}2.09 & \textbf{43.02\hspace{0.02em}$\pm$\hspace{0.02em}1.30} & 51.80\hspace{0.02em}$\pm$\hspace{0.02em}0.82 & \textbf{77.91\hspace{0.02em}$\pm$\hspace{0.02em}0.71} & \textbf{76.64\hspace{0.02em}$\pm$\hspace{0.02em}1.25} & \textbf{78.66\hspace{0.02em}$\pm$\hspace{0.02em}0.55} & \textbf{81.06\hspace{0.02em}$\pm$\hspace{0.02em}0.78} \\
    
    Inception-v3 & 25.00\hspace{0.02em}$\pm$\hspace{0.02em}3.91 & 37.26\hspace{0.02em}$\pm$\hspace{0.02em}1.23 & 42.06\hspace{0.02em}$\pm$\hspace{0.02em}0.69 & 51.67\hspace{0.02em}$\pm$\hspace{0.02em}1.20 & 75.15\hspace{0.02em}$\pm$\hspace{0.02em}1.09 & 73.69\hspace{0.02em}$\pm$\hspace{0.02em}1.42 & 75.49\hspace{0.02em}$\pm$\hspace{0.02em}0.91 & 78.33\hspace{0.02em}$\pm$\hspace{0.02em}0.40 \\
    ResNet-18 & \textbf{29.01\hspace{0.02em}$\pm$\hspace{0.02em}3.63} & 37.54\hspace{0.02em}$\pm$\hspace{0.02em}0.62 & 42.78\hspace{0.02em}$\pm$\hspace{0.02em}1.30 & 51.50\hspace{0.02em}$\pm$\hspace{0.02em}1.37 & 75.29\hspace{0.02em}$\pm$\hspace{0.02em}1.05 & 73.94\hspace{0.02em}$\pm$\hspace{0.02em}1.11 & 76.65\hspace{0.02em}$\pm$\hspace{0.02em}1.48 & 79.13\hspace{0.02em}$\pm$\hspace{0.02em}0.75 \\
    WRN-16-8 & 22.64\hspace{0.02em}$\pm$\hspace{0.02em}3.82 & \textbf{37.71\hspace{0.02em}$\pm$\hspace{0.02em}1.73} & 40.78\hspace{0.02em}$\pm$\hspace{0.02em}1.79 & \textbf{53.02\hspace{0.02em}$\pm$\hspace{0.02em}1.80} & 76.64\hspace{0.02em}$\pm$\hspace{0.02em}0.92 & 75.84\hspace{0.02em}$\pm$\hspace{0.02em}0.84 & 77.19\hspace{0.02em}$\pm$\hspace{0.02em}1.14 & 80.77\hspace{0.02em}$\pm$\hspace{0.02em}0.30 \\
    \bottomrule
    \end{tabular}%
    }
  \label{tab:cross}%
\end{table}%

\subsection{Sensitiveness to Pre-trained Models}
As previously mentioned, some coreset selection methods rely on a pre-trained model to obtain metrics, \eg feature, gradient and loss, for selecting samples.
This experiment explores the influence of the pre-trained models, which are pre-trained for different epochs, on the final coreset performance.
Similar to Sec. \ref{ssec:cross}, four representative methods (\textsc{Forgetting}, \textsc{Glister}, \textsc{GraNd} and \textsc{Graph Cut}) and two selection fractions (1\% and 10\%) are tested in this experiment.
Except for different pre-training epochs, all other settings and hyper-parameters are consistent with those in Sec. \ref{ssec:cifar}.
We report our results in Tab. \ref{tab:pretrain}.
For \textsc{Forgetting}, good results can be achieved with models pre-trained for only 2 epochs, \ie selecting samples based on whether the first forgetting event occurs on each sample. Spending more epochs in calculating forgetting events does not lead to improvements. 
The forgetting events can only be counted for more than 2 training epochs, thus no results are provided for \textsc{Forgetting} in epoch 0 and 1.
\textsc{GraNd} also performs best with models pre-trained for 2 epochs.
The results indicate that it is not necessary to pre-train a model for too many epochs to obtain the metrics.

\begin{table}[htbp]
\setlength\tabcolsep{3pt}
\centering
  \caption{Sensitiveness to pre-trained models. Performance (\%) of different methods using pre-trained models with varying pre-training epochs.}  \label{tab:pretrain}
  \resizebox{1\linewidth}{!}
  {
    \begin{tabular}{cccccccccccc}
    \toprule
    Pre-train Epochs & 0     & 1     & 2     & 5     & 10    & 15    & 20    & 50    & 100   & 150   & 200 \\
    \midrule
          & \multicolumn{11}{c}{\textbf{1\%}} \\  \cmidrule(lr){2-12}\cmidrule(lr){1-1}
    \textbf{Forgetting} & - & - & 36.06\hspace{0.02em}$\pm$\hspace{0.02em}0.65 & 36.81\hspace{0.02em}$\pm$\hspace{0.02em}1.82 & 35.20\hspace{0.02em}$\pm$\hspace{0.02em}1.59 & 32.96\hspace{0.02em}$\pm$\hspace{0.02em}1.20 & 32.22\hspace{0.02em}$\pm$\hspace{0.02em}1.01 & 24.23\hspace{0.02em}$\pm$\hspace{0.02em}0.64 & 20.41\hspace{0.02em}$\pm$\hspace{0.02em}0.91 & 19.84\hspace{0.02em}$\pm$\hspace{0.02em}0.56 & 19.47\hspace{0.02em}$\pm$\hspace{0.02em}0.30 \\
    \textbf{GraNd} & 28.17\hspace{0.02em}$\pm$\hspace{0.02em}0.20  & 31.05\hspace{0.02em}$\pm$\hspace{0.02em}1.36 & 31.24\hspace{0.02em}$\pm$\hspace{0.02em}2.36 & 29.70\hspace{0.02em}$\pm$\hspace{0.02em}1.02  & 26.69\hspace{0.02em}$\pm$\hspace{0.02em}1.30 & 26.11\hspace{0.02em}$\pm$\hspace{0.02em}1.46 & 26.39\hspace{0.02em}$\pm$\hspace{0.02em}0.89 & 26.81\hspace{0.02em}$\pm$\hspace{0.02em}1.97 & 26.52\hspace{0.02em}$\pm$\hspace{0.02em}1.10  & 26.08\hspace{0.02em}$\pm$\hspace{0.02em}0.65 & 27.17\hspace{0.02em}$\pm$\hspace{0.02em}1.84 \\
    \textbf{Glister} & 27.63\hspace{0.02em}$\pm$\hspace{0.02em}0.85 & 33.97\hspace{0.02em}$\pm$\hspace{0.02em}2.68 & 33.31\hspace{0.02em}$\pm$\hspace{0.02em}1.08 & 32.93\hspace{0.02em}$\pm$\hspace{0.02em}1.51 & 32.93\hspace{0.02em}$\pm$\hspace{0.02em}2.36 & 32.28\hspace{0.02em}$\pm$\hspace{0.02em}2.09 & 31.15\hspace{0.02em}$\pm$\hspace{0.02em}2.24 & 31.46\hspace{0.02em}$\pm$\hspace{0.02em}1.56 & 32.89\hspace{0.02em}$\pm$\hspace{0.02em}1.24 & 33.37\hspace{0.02em}$\pm$\hspace{0.02em}1.91 & 34.06\hspace{0.02em}$\pm$\hspace{0.02em}2.17 \\
    \textbf{Graph Cut} & 33.61\hspace{0.02em}$\pm$\hspace{0.02em}1.40 & 43.15\hspace{0.02em}$\pm$\hspace{0.02em}1.31 & 43.00\hspace{0.02em}$\pm$\hspace{0.02em}0.76 & 44.33\hspace{0.02em}$\pm$\hspace{0.02em}1.55 & 42.78\hspace{0.02em}$\pm$\hspace{0.02em}1.30 & 41.33\hspace{0.02em}$\pm$\hspace{0.02em}2.01 & 41.30\hspace{0.02em}$\pm$\hspace{0.02em}2.80 & 42.23\hspace{0.02em}$\pm$\hspace{0.02em}1.72 & 40.46\hspace{0.02em}$\pm$\hspace{0.02em}0.93 & 41.74\hspace{0.02em}$\pm$\hspace{0.02em}1.46 & 40.53\hspace{0.02em}$\pm$\hspace{0.02em}2.27 \\
    \midrule
          & \multicolumn{11}{c}{\textbf{10\%}} \\
          \cmidrule(lr){2-12}\cmidrule(lr){1-1}
    \textbf{Forgetting} & - & - & 72.62\hspace{0.02em}$\pm$\hspace{0.02em}2.79 & 72.72\hspace{0.02em}$\pm$\hspace{0.02em}1.44 & 66.99\hspace{0.02em}$\pm$\hspace{0.02em}1.48 & 60.87\hspace{0.02em}$\pm$\hspace{0.02em}1.92 & 54.62\hspace{0.02em}$\pm$\hspace{0.02em}2.48 & 44.10\hspace{0.02em}$\pm$\hspace{0.02em}1.21 & 42.29\hspace{0.02em}$\pm$\hspace{0.02em}1.01 & 41.97\hspace{0.02em}$\pm$\hspace{0.02em}0.70 & 41.99\hspace{0.02em}$\pm$\hspace{0.02em}1.02 \\
    \textbf{GraNd} & 62.54\hspace{0.02em}$\pm$\hspace{0.02em}2.15 & 63.15\hspace{0.02em}$\pm$\hspace{0.02em}1.99 & 71.34\hspace{0.02em}$\pm$\hspace{0.02em}1.82 & 67.97\hspace{0.02em}$\pm$\hspace{0.02em}1.86 & 52.73\hspace{0.02em}$\pm$\hspace{0.02em}1.86 & 64.76\hspace{0.02em}$\pm$\hspace{0.02em}1.83 & 65.20\hspace{0.02em}$\pm$\hspace{0.02em}1.21  & 66.33\hspace{0.02em}$\pm$\hspace{0.02em}2.29 & 57.21\hspace{0.02em}$\pm$\hspace{0.02em}1.75 & 58.36\hspace{0.02em}$\pm$\hspace{0.02em}1.49 & 65.34\hspace{0.02em}$\pm$\hspace{0.02em}0.55 \\
    \textbf{Glister} & 59.35\hspace{0.02em}$\pm$\hspace{0.02em}2.31 & 60.83\hspace{0.02em}$\pm$\hspace{0.02em}3.18 & 68.79\hspace{0.02em}$\pm$\hspace{0.02em}1.15 & 68.81\hspace{0.02em}$\pm$\hspace{0.02em}2.75 & 66.26\hspace{0.02em}$\pm$\hspace{0.02em}3.47 & 61.99\hspace{0.02em}$\pm$\hspace{0.02em}3.05 & 68.03\hspace{0.02em}$\pm$\hspace{0.02em}1.72 & 65.05\hspace{0.02em}$\pm$\hspace{0.02em}1.66 & 66.26\hspace{0.02em}$\pm$\hspace{0.02em}2.92 & 68.16\hspace{0.02em}$\pm$\hspace{0.02em}2.78 & 68.16\hspace{0.02em}$\pm$\hspace{0.02em}3.03 \\
    \textbf{Graph Cut} & 63.39\hspace{0.02em}$\pm$\hspace{0.02em}1.54 & 62.52\hspace{0.02em}$\pm$\hspace{0.02em}1.02 & 68.26\hspace{0.02em}$\pm$\hspace{0.02em}1.11 & 72.91\hspace{0.02em}$\pm$\hspace{0.02em}1.13 & 76.65\hspace{0.02em}$\pm$\hspace{0.02em}1.48 & 77.06\hspace{0.02em}$\pm$\hspace{0.02em}1.09 & 68.73\hspace{0.02em}$\pm$\hspace{0.02em}0.87 & 77.48\hspace{0.02em}$\pm$\hspace{0.02em}0.51 & 76.66\hspace{0.02em}$\pm$\hspace{0.02em}1.64 & 76.16\hspace{0.02em}$\pm$\hspace{0.02em}2.14 & 76.33\hspace{0.02em}$\pm$\hspace{0.02em}1.52 \\
    \bottomrule
    \end{tabular}%
}
\end{table}%



\section{Extended Related Work}
\label{sec:relatedwork}
An alternative way to reduce training set size is dataset condensation (or distillation) \cite{wang2018dataset,zhao2021DC,zhao2021DSA}. Instead of selecting subsets, it learns to synthesize informative training samples that can be more informative than real samples in the original training set. 
Although remarkable progress has been achieved in this research area, it is still challenging to apply dataset condensation on large-scale and high-resolution datasets, \eg ImageNet-1K, due to the expensive and difficult optimization.

\section{Conclusion}
\label{sec:conclusion}
In this work, we contribute a comprehensive code library -- \textit{DeepCore} for coreset selection in deep learning, where we re-implement dozens of state-of-the-art coreset selection methods on popular datasets and network architectures. Our code library enables a convenient and fair comparison of methods in various learning settings. Extensive experiments on CIFAR10 and ImageNet datasets verify that, although various methods have advantages in certain experiment settings, random selection is still a strong baseline.

\subsubsection*{Acknowledgment.} This research was supported by Public Health \& Disease Control and Prevention, Major Innovation \& Planning Interdisciplinary Platform for the ``Double-First Class'' Initiative, Renmin University of China (No. 2022PDPC), fund for building world-class universities (disciplines) of Renmin University of China. Project No. KYGJA2022001. This research was supported by Public Computing Cloud, Renmin University of China.

%
%
%
\newpage
\bibliographystyle{splncs04}
\bibliography{refs}

\begin{thebibliography}{10}
\providecommand{\url}[1]{\texttt{#1}}
\providecommand{\urlprefix}{URL }
\providecommand{\doi}[1]{https://doi.org/#1}

\bibitem{agarwal2020contextual}
Agarwal, S., Arora, H., Anand, S., Arora, C.: Contextual diversity for active
  learning. In: ECCV. pp. 137--153. Springer (2020)

\bibitem{aljundi2019gradient}
Aljundi, R., Lin, M., Goujaud, B., Bengio, Y.: Gradient based sample selection
  for online continual learning. Advances in Neural Information Processing
  Systems  \textbf{32},  11816--11825 (2019)

\bibitem{bachem2015coresets}
Bachem, O., Lucic, M., Krause, A.: Coresets for nonparametric estimation-the
  case of dp-means. In: ICML. pp. 209--217. PMLR (2015)

\bibitem{bateni2014distributed}
Bateni, M., Bhaskara, A., Lattanzi, S., Mirrokni, V.S.: Distributed balanced
  clustering via mapping coresets. In: NIPS. pp. 2591--2599 (2014)

\bibitem{borsos2020coresets}
Borsos, Z., Mutny, M., Krause, A.: Coresets via bilevel optimization for
  continual learning and streaming. Advances in Neural Information Processing
  Systems  \textbf{33} (2020)

\bibitem{borsos2021semi}
Borsos, Z., Tagliasacchi, M., Krause, A.: Semi-supervised batch active learning
  via bilevel optimization. In: ICASSP 2021. pp. 3495--3499. IEEE (2021)

\bibitem{chen2010super}
Chen, Y., Welling, M., Smola, A.: Super-samples from kernel herding. The
  Twenty-Sixth Conference Annual Conference on Uncertainty in Artificial
  Intelligence  (2010)

\bibitem{chhaya2020coresets}
Chhaya, R., Dasgupta, A., Shit, S.: On coresets for regularized regression. In:
  International Conference on Machine Learning. pp. 1866--1876. PMLR (2020)

\bibitem{coleman2019selection}
Coleman, C., Yeh, C., Mussmann, S., Mirzasoleiman, B., Bailis, P., Liang, P.,
  Leskovec, J., Zaharia, M.: Selection via proxy: Efficient data selection for
  deep learning. In: ICLR (2019)

\bibitem{dasgupta2019teaching}
Dasgupta, S., Hsu, D., Poulis, S., Zhu, X.: Teaching a black-box learner. In:
  ICML. PMLR (2019)

\bibitem{ducoffe2018adversarial}
Ducoffe, M., Precioso, F.: Adversarial active learning for deep networks: a
  margin based approach. arXiv preprint arXiv:1802.09841  (2018)

\bibitem{wolf2011facility}
Farahani, R.Z., Hekmatfar, M.: Facility location: concepts, models, algorithms
  and case studies (2009)

\bibitem{feldman2011scalable}
Feldman, D., Faulkner, M., Krause, A.: Scalable training of mixture models via
  coresets. In: NIPS. pp. 2142--2150. Citeseer (2011)

\bibitem{he2016deep}
He, K., Zhang, X., Ren, S., Sun, J.: Deep residual learning for image
  recognition. In: Proceedings of the IEEE conference on computer vision and
  pattern recognition. pp. 770--778 (2016)

\bibitem{DBLP:journals/corr/abs-1905-02244}
Howard, A., Sandler, M., Chu, G., Chen, L., Chen, B., Tan, M., Wang, W., Zhu,
  Y., Pang, R., Vasudevan, V., Le, Q.V., Adam, H.: Searching for mobilenetv3
  (2019), \url{http://arxiv.org/abs/1905.02244}

\bibitem{iyer2021submodular}
Iyer, R., Khargoankar, N., Bilmes, J., Asanani, H.: Submodular combinatorial
  information measures with applications in machine learning. In: Algorithmic
  Learning Theory. pp. 722--754. PMLR (2021)

\bibitem{iyer2013submodular}
Iyer, R.K., Bilmes, J.A.: Submodular optimization with submodular cover and
  submodular knapsack constraints. Advances in neural information processing
  systems  \textbf{26} (2013)

\bibitem{ju2021extending}
Ju, J., Jung, H., Oh, Y., Kim, J.: Extending contrastive learning to
  unsupervised coreset selection. arXiv preprint arXiv:2103.03574  (2021)

\bibitem{kaushal2021prism}
Kaushal, V., Kothawade, S., Ramakrishnan, G., Bilmes, J., Iyer, R.: Prism: A
  unified framework of parameterized submodular information measures for
  targeted data subset selection and summarization. arXiv preprint
  arXiv:2103.00128  (2021)

\bibitem{pmlr-v139-killamsetty21a}
Killamsetty, K., Durga, S., Ramakrishnan, G., De, A., Iyer, R.: Grad-match:
  Gradient matching based data subset selection for efficient deep model
  training. In: ICML. pp. 5464--5474 (2021)

\bibitem{killamsetty2021glister}
Killamsetty, K., Sivasubramanian, D., Ramakrishnan, G., Iyer, R.: Glister:
  Generalization based data subset selection for efficient and robust learning.
  In: Proceedings of the AAAI Conference on Artificial Intelligence (2021)

\bibitem{killamsetty2021retrieve}
Killamsetty, K., Zhao, X., Chen, F., Iyer, R.: Retrieve: Coreset selection for
  efficient and robust semi-supervised learning. arXiv preprint
  arXiv:2106.07760  (2021)

\bibitem{knoblauch2020optimal}
Knoblauch, J., Husain, H., Diethe, T.: Optimal continual learning has perfect
  memory and is np-hard. In: International Conference on Machine Learning. pp.
  5327--5337. PMLR (2020)

\bibitem{kothawade2021similar}
Kothawade, S., Beck, N., Killamsetty, K., Iyer, R.: Similar: Submodular
  information measures based active learning in realistic scenarios. arXiv
  preprint arXiv:2107.00717  (2021)

\bibitem{krizhevsky2009learning}
Krizhevsky, A., Hinton, G., et~al.: Learning multiple layers of features from
  tiny images  (2009)

\bibitem{NIPS2012_c399862d}
Krizhevsky, A., Sutskever, I., Hinton, G.E.: Imagenet classification with deep
  convolutional neural networks. In: Pereira, F., Burges, C.J.C., Bottou, L.,
  Weinberger, K.Q. (eds.) Advances in Neural Information Processing Systems.
  vol.~25. Curran Associates, Inc. (2012)

\bibitem{le2015tiny}
Le, Y., Yang, X.: Tiny imagenet visual recognition challenge. CS 231N
  \textbf{7}(7), ~3 (2015)

\bibitem{lecun1989backpropagation}
LeCun, Y., Boser, B., Denker, J.S., Henderson, D., Howard, R.E., Hubbard, W.,
  Jackel, L.D.: Backpropagation applied to handwritten zip code recognition.
  Neural computation  \textbf{1}(4),  541--551 (1989)

\bibitem{lecun1998gradient}
LeCun, Y., Bottou, L., Bengio, Y., Haffner, P., et~al.: Gradient-based learning
  applied to document recognition. Proceedings of the IEEE  \textbf{86}(11),
  2278--2324 (1998)

\bibitem{liu2021just}
Liu, E.Z., Haghgoo, B., Chen, A.S., Raghunathan, A., Koh, P.W., Sagawa, S.,
  Liang, P., Finn, C.: Just train twice: Improving group robustness without
  training group information. In: ICML. pp. 6781--6792 (2021)

\bibitem{DBLP:journals/corr/abs-2109-03764}
Margatina, K., Vernikos, G., Barrault, L., Aletras, N.: Active learning by
  acquiring contrastive examples. arXiv preprint arXiv:2109.03764  (2021)

\bibitem{mirzasoleiman2020coresets}
Mirzasoleiman, B., Bilmes, J., Leskovec, J.: Coresets for data-efficient
  training of machine learning models. In: ICML. PMLR (2020)

\bibitem{NEURIPS2020_8493eeac}
Mirzasoleiman, B., Cao, K., Leskovec, J.: Coresets for robust training of deep
  neural networks against noisy labels (2020)

\bibitem{munteanu2018coresets}
Munteanu, A., Schwiegelshohn, C., Sohler, C., Woodruff, D.P.: On coresets for
  logistic regression. In: NeurIPS (2018)

\bibitem{nemhauser1978analysis}
Nemhauser, G.L., Wolsey, L.A., Fisher, M.L.: An analysis of approximations for
  maximizing submodular set functions—i. Mathematical programming
  \textbf{14}(1),  265--294 (1978)

\bibitem{netzer2011reading}
Netzer, Y., Wang, T., Coates, A., Bissacco, A., Wu, B., Ng, A.Y.: Reading
  digits in natural images with unsupervised feature learning  (2011)

\bibitem{paszke2019pytorch}
Paszke, A., Gross, S., Massa, F., Lerer, A., Bradbury, J., Chanan, G., Killeen,
  T., Lin, Z., Gimelshein, N., Antiga, L., et~al.: Pytorch: An imperative
  style, high-performance deep learning library. Advances in neural information
  processing systems  \textbf{32} (2019)

\bibitem{paul2021deep}
Paul, M., Ganguli, S., Dziugaite, G.K.: Deep learning on a data diet: Finding
  important examples early in training. arXiv preprint arXiv:2107.07075  (2021)

\bibitem{ILSVRC15}
Russakovsky, O., Deng, J., Su, H., Krause, J., Satheesh, S., Ma, S., Huang, Z.,
  Karpathy, A., Khosla, A., Bernstein, M., Berg, A.C., Fei-Fei, L.: {ImageNet
  Large Scale Visual Recognition Challenge}. IJCV  (2015)

\bibitem{sachdeva2021svp}
Sachdeva, N., Wu, C.J., McAuley, J.: Svp-cf: Selection via proxy for
  collaborative filtering data. arXiv preprint arXiv:2107.04984  (2021)

\bibitem{sener2018active}
Sener, O., Savarese, S.: Active learning for convolutional neural networks: A
  core-set approach. In: ICLR (2018)

\bibitem{settles2009active}
Settles, B.: Active learning literature survey  (2009)

\bibitem{settles2011theories}
Settles, B.: From theories to queries: Active learning in practice. In: Active
  learning and experimental design workshop in conjunction with AISTATS 2010.
  pp. 1--18. JMLR Workshop and Conference Proceedings (2011)

\bibitem{shim2021core}
Shim, J.h., Kong, K., Kang, S.J.: Core-set sampling for efficient neural
  architecture search. arXiv preprint arXiv:2107.06869  (2021)

\bibitem{simonyan2014very}
Simonyan, K., Zisserman, A.: Very deep convolutional networks for large-scale
  image recognition. arXiv preprint arXiv:1409.1556  (2014)

\bibitem{sinha2020small}
Sinha, S., Zhang, H., Goyal, A., Bengio, Y., Larochelle, H., Odena, A.:
  Small-gan: Speeding up gan training using core-sets. In: ICML. PMLR (2020)

\bibitem{sohler2018strong}
Sohler, C., Woodruff, D.P.: Strong coresets for k-median and subspace
  approximation: Goodbye dimension. In: 2018 IEEE 59th Annual Symposium on
  Foundations of Computer Science (FOCS). pp. 802--813. IEEE (2018)

\bibitem{szegedy2016rethinking}
Szegedy, C., Vanhoucke, V., Ioffe, S., Shlens, J., Wojna, Z.: Rethinking the
  inception architecture for computer vision. In: Proceedings of the IEEE
  conference on computer vision and pattern recognition. pp. 2818--2826 (2016)

\bibitem{toneva2018empirical}
Toneva, M., Sordoni, A., des Combes, R.T., Trischler, A., Bengio, Y., Gordon,
  G.J.: An empirical study of example forgetting during deep neural network
  learning. In: ICLR (2018)

\bibitem{wang2018dataset}
Wang, T., Zhu, J.Y., Torralba, A., Efros, A.A.: Dataset distillation. arXiv
  preprint arXiv:1811.10959  (2018)

\bibitem{wei2015submodularity}
Wei, K., Iyer, R., Bilmes, J.: Submodularity in data subset selection and
  active learning. In: International Conference on Machine Learning. PMLR
  (2015)

\bibitem{welling2009herding}
Welling, M.: Herding dynamical weights to learn. In: Proceedings of the 26th
  Annual International Conference on Machine Learning. pp. 1121--1128 (2009)

\bibitem{xiao2017fashion}
Xiao, H., Rasul, K., Vollgraf, R.: Fashion-mnist: a novel image dataset for
  benchmarking machine learning algorithms. arXiv preprint arXiv:1708.07747
  (2017)

\bibitem{yadav2019cold}
Yadav, C., Bottou, L.: Cold case: The lost mnist digits. Advances in neural
  information processing systems  \textbf{32} (2019)

\bibitem{yoon2021online}
Yoon, J., Madaan, D., Yang, E., Hwang, S.J.: Online coreset selection for
  rehearsal-based continual learning. arXiv preprint arXiv:2106.01085  (2021)

\bibitem{zagoruyko2016wide}
Zagoruyko, S., Komodakis, N.: Wide residual networks. arXiv preprint
  arXiv:1605.07146  (2016)

\bibitem{zhao2021DSA}
Zhao, B., Bilen, H.: Dataset condensation with differentiable siamese
  augmentation. In: International Conference on Machine Learning (2021)

\bibitem{zhao2021DC}
Zhao, B., Mopuri, K.R., Bilen, H.: Dataset condensation with gradient matching.
  In: International Conference on Learning Representations (2021),
  \url{https://openreview.net/forum?id=mSAKhLYLSsl}

\end{thebibliography}

\end{document}